\pdfoutput=1
\documentclass[11pt]{article}

\usepackage[preprint]{acl}

\usepackage{times}
\usepackage{latexsym}

\usepackage[T1]{fontenc}
\usepackage[utf8]{inputenc}

\usepackage{microtype}

\usepackage{inconsolata}

\usepackage{graphicx}
\usepackage{tabularx}
\usepackage{amssymb}
\usepackage{amsmath}
\usepackage{python}
\usepackage{pythontex}
\usepackage{algorithm}
\usepackage{algpseudocode}
\usepackage{arydshln}
\usepackage{booktabs}
\usepackage{CJKutf8}
\usepackage{multicol}

\title{M-Ped: Multi-Prompt Ensemble Decoding for Large Language Models}

\author{Jiaxin GUO, Daimeng Wei, Yuanchang Luo, Shimin Tao, Hengchao Shang, Zongyao Li\\
    \textbf{Shaojun Li, Jinlong Yang, Zhanglin Wu, Zhiqiang Rao and Hao Yang} \\
        \{jiaxinguo1,weidaimeng,luoyuanchang1,taoshimin,shanghengchao,lizongyao\}@huawei.com\\
        \{lishaojun18,yangjinlong7,wuzhanglin2,raozhiqiang,yanghao30\}@huawei.com\\
        Huawei Translation Services Center, Beijing, China }

\begin{document}
\maketitle
\begin{abstract}
% 关键词： Prompt、 Ensemble、 LLMs
% With the widespread application of Large Language Models (LLMs) in the field of Natural Language Processing (NLP), enhancing their performance has become a research hotspot. This study introduces a multi-prompt ensemble decoding method aimed at improving the generation quality, diversity, and efficiency of LLMs by integrating the results of multiple prompts. Our method encompasses two core technologies: Efficient Batch Processing with Left-Padding and Uniform Averaging for Prompt Ensemble. The Efficient Batch Processing with Left-Padding technique standardizes the length of all prompts through preprocessing to meet the model's batch processing requirements, while the Uniform Averaging for Prompt Ensemble technique reduces bias and enhances model robustness by uniformly averaging the outputs from multiple prompts. We conducted extensive experiments across various tasks, including machine translation, code generation, and text simplification, demonstrating the effectiveness of our method in enhancing LLMs' performance. The experimental results indicate that our approach achieved significant improvements in BLEU, pass@$k$ pass rates, and LENS metrics compared to the original methods. This study not only provides a new perspective for the application of LLMs but also contributes significant technological advancements to the development of the NLP field.

With the widespread application of Large Language Models (LLMs) in the field of Natural Language Processing (NLP), enhancing their performance has become a research hotspot. This paper presents a novel multi-prompt ensemble decoding approach designed to bolster the generation quality of LLMs by leveraging the aggregation of outcomes from multiple prompts.
Given a unique input $X$, we submit $n$ variations of prompts with $X$ to LLMs in batch mode to decode and derive probability distributions. For each token prediction, we calculate the ensemble probability by averaging the $n$ probability distributions within the batch, utilizing this aggregated probability to generate the token. This technique is dubbed Inner-Batch Ensemble. To facilitate efficient batch inference, we implement a Left-Padding strategy to maintain uniform input lengths across the n prompts.
Through extensive experimentation on diverse NLP tasks, including machine translation, code generation, and text simplification, we demonstrate the efficacy of our method in enhancing LLM performance. The results show substantial improvements in BLEU scores, pass@$k$ rates, and LENS metrics over conventional methods. 
\end{abstract}

\section{Introduction}

% 关键词： Prompt、 Ensemble、 LLMs

\begin{figure}[th]
\centering
\includegraphics[width=7.5cm]{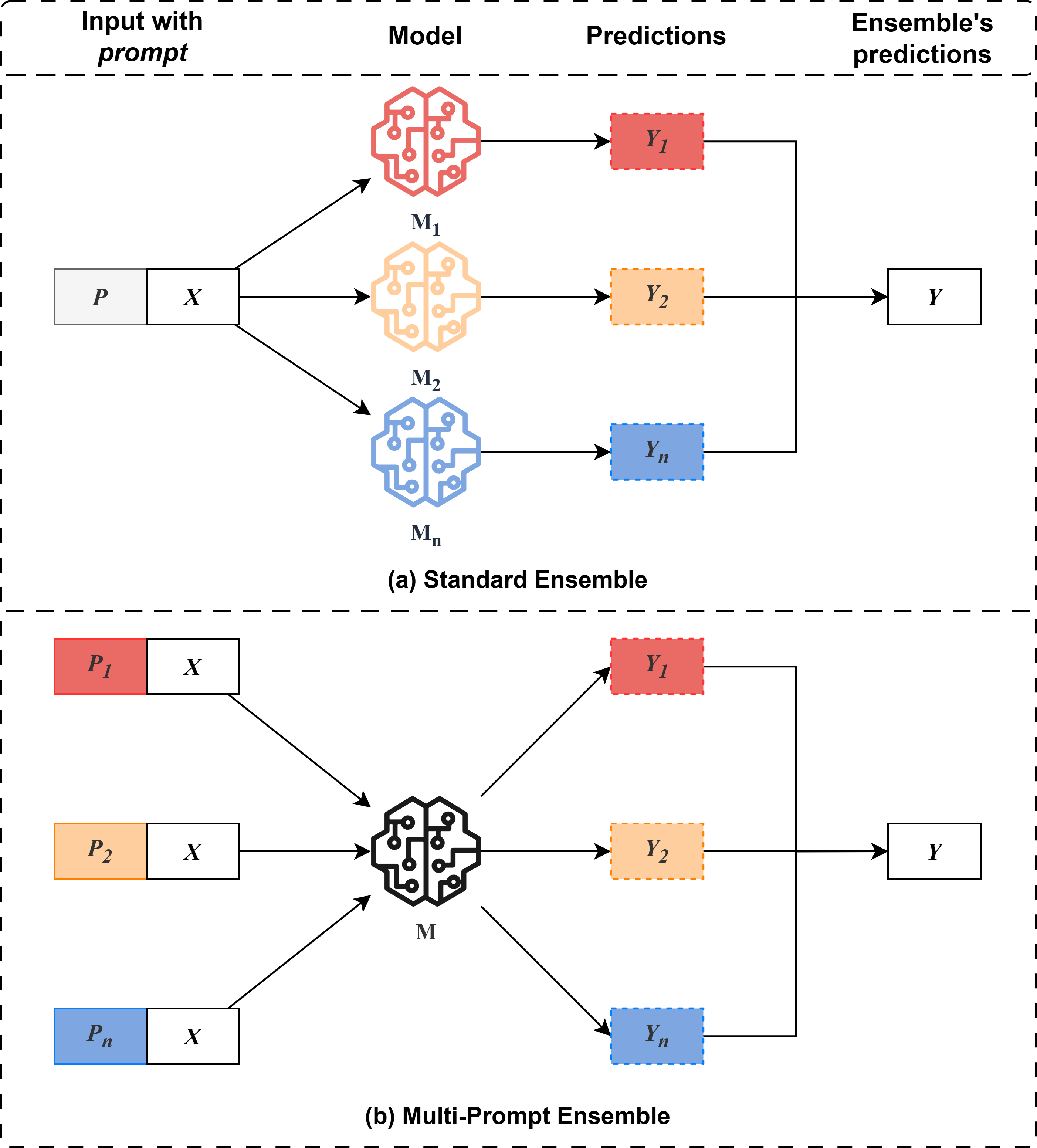}
\caption{(a) \textbf{Standard Ensemble}: Input $X$ with prompt $P$ is fed into models $M_1$, $M_2$, ..., $M_n$, yielding predictions $Y_1$, $Y_2$, ..., $Y_n$. These are combined to get the final ensemble result $Y$. (b) \textbf{Our Multi-Prompt Ensemble}: Input $X$ constructs prompts $P_1$, $P_2$, ..., $P_n$, forming diverse samples batched through one model $M$ to get predictions $Y_1$, $Y_2$, ..., $Y_n$. The ensemble result $Y$ is the average of these predictions.}
% \caption{(a) Standard Ensemble: A single input $x$ with prompt $p$ is passed through distinct models $M\_1$, $M\_2$, ..., $M\_n$, each generating its own prediction $y\_1$, $y\_2$, ..., $y\_n$. These individual predictions are then combined to produce the final ensemble result $y$. (b) Our Multi-Prompt Ensemble: A single input $x$ is used to construct multiple prompts $p\_1$, $p\_2$, ..., $p\_n$, creating diverse input samples. These samples are batched and passed through a single model $M$, yielding a set of predictions $y\_1$, $y\_2$, ..., $y\_n$. An ensemble result $y$ is obtained by averaging the prediction probabilities within the batch.}
\label{figure:m_ped_1}
\end{figure}

\begin{figure*}[th]
\centering
\includegraphics[width=15.6cm]{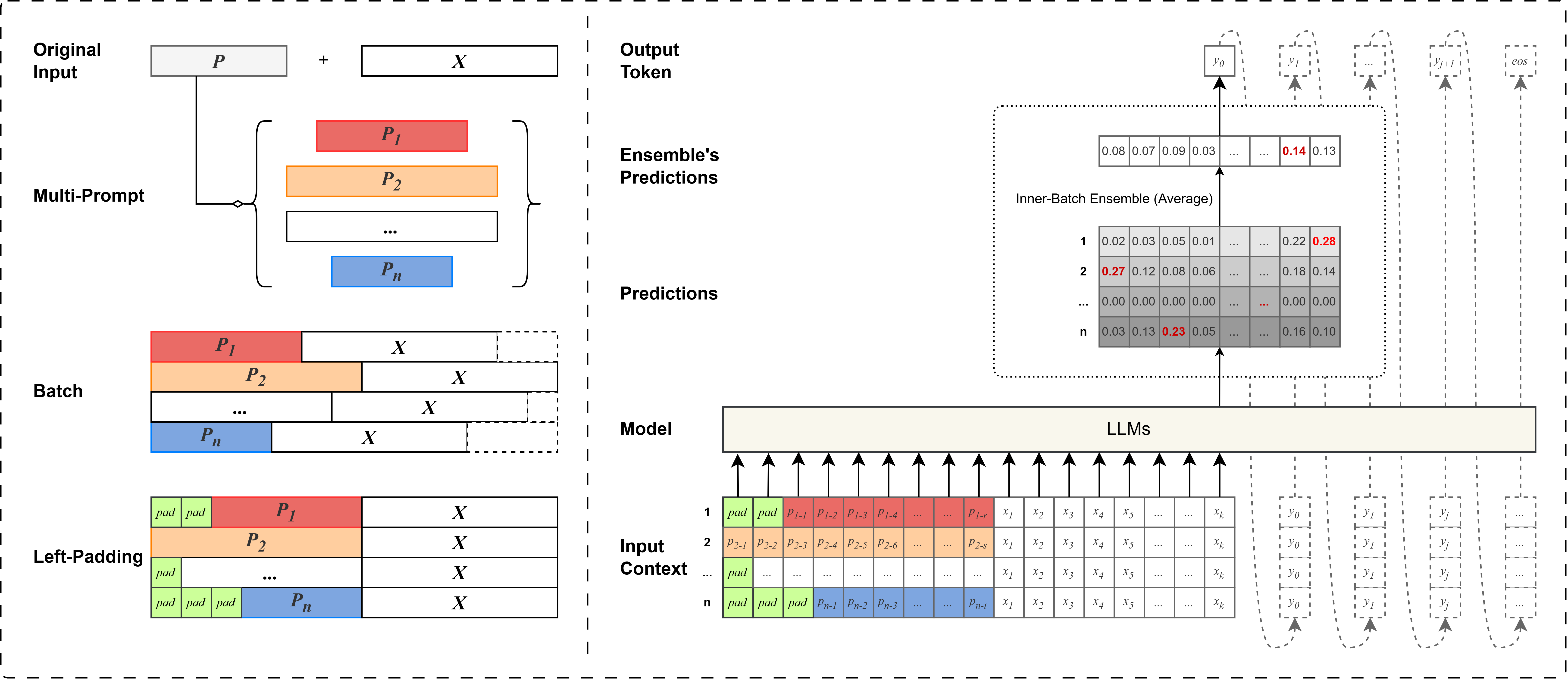}
\caption{The overall process of Our Multi-Prompt Ensemble Decoding.}
\label{figure:model}
\end{figure*}

% 大模型 + Prompt
Large Language Models (LLMs) \cite{DBLP:journals/corr/abs-2303-08774,DBLP:journals/corr/abs-2302-13971,DBLP:journals/corr/abs-2307-09288,DBLP:journals/corr/abs-2309-16609,DBLP:journals/corr/abs-2407-10671} have demonstrated exceptional capabilities in understanding and generating natural language through extensive data pre-training, becoming the core driving force in the field of Natural Language Processing (NLP). Prompt technology \cite{DBLP:journals/tacl/JiangXAN20,DBLP:journals/csur/LiuYFJHN23,DBLP:journals/corr/abs-2304-05970,DBLP:conf/ijcai/ZhaoWY23,DBLP:conf/emnlp/HeinemanD024,DBLP:conf/nips/Wei0SBIXCLZ22,DBLP:conf/iclr/0002WSLCNCZ23}, as a key to enhancing LLMs' performance, can strengthen the model's effects without altering model parameters, achieving seamless integration with downstream tasks. However, in the practical application of LLMs, the output results are closely related to the quality of the Prompt, and a precise and effective Prompt is crucial for improving the model's response quality. 
% Therefore, the design and optimization of Prompts have become key factors in enhancing LLMs' performance and an important trend in the development of large model technology. 
How to use Prompts more efficiently to fully leverage the potential of LLMs has become a hot issue of common concern in academia and industry.

Ensemble decoding \cite{DBLP:conf/nips/Lakshminarayanan17,DBLP:journals/eaai/GanaieHMTS22,DBLP:journals/ai/ZhouWT02} is a widely employed technique for enhancing the quality of model-generated outputs. Typically, as shown in Figure \ref{figure:m_ped_1}(a), standard ensemble decoding refers to the process of combining the outputs of multiple distinct models on the same input. Specifically, during the prediction of each word, ensemble decoding generates a comprehensive prediction by averaging the probability distributions provided by multiple models. This approach leverages the diversity among models, aiming to reduce uncertainty and improve overall predictive performance. In theory, we could apply ensemble decoding to LLMs. However, LLMs already demand considerable memory resources, and the implementation of ensemble decoding with multiple LLMs presents a significant challenge in terms of memory usage. Moreover, even if these models could be deployed, optimizing the performance of the ensemble decoding becomes a complex issue.

% This paper
% This study aims to address the aforementioned challenges by proposing a multi-prompt ensemble decoding method. The method includes two key technologies: Efficient Batch Processing with Left-Padding and Uniform Averaging for Prompt Ensemble. The Efficient Batch Processing with Left-Padding technique standardizes the length of all prompts by padding special characters in front of shorter ones before they are input into the model, thus adapting to the model's batch processing requirements, improving processing efficiency, and allowing the model to make more effective use of its internal parallel computing resources. The Uniform Averaging for Prompt Ensemble technique reduces the bias that may be introduced by a single prompt and enhances the model's robustness by uniformly averaging the outputs from multiple prompts. We have demonstrated the effectiveness of this method in improving the generation quality, diversity, and efficiency of LLMs across various tasks, including machine translation, code generation, and text simplification.

% Methods
In this paper, we introduce a particularly simple yet effective method: Multi-Prompt Ensemble Decoding (M-Ped). This approach constructs $n$ distinct prompts for a single query, generating $n$ diverse input samples that are batched together and submitted to LLMs for inference. During the inference process, we average the prediction probabilities within the batch for each word prediction. To ensure batched inference is feasible, we specifically propose the use of left-padding technology to address the issue of varying prompt lengths within a batch. Compared to traditional ensemble methods, our approach shifts the focus of diversity from using different models to using different prompts. We have validated the effectiveness of this method across various tasks and multiple models, including extensive experiments on multiple test sets for machine translation, code generation, and text simplification tasks. The results demonstrate improvements in BLEU scores for machine translation\cite{DBLP:conf/acl/PapineniRWZ02,DBLP:conf/nips/VaswaniSPUJGKP17,sennrich2016improving,wei2023text,DBLP:conf/iclr/Gu0XLS18}, pass@$k$ scores for code generation\cite{DBLP:journals/corr/abs-2107-03374,DBLP:journals/corr/abs-2406-00515,DBLP:journals/access/DehaerneDHGM22}, and LENS scores for text simplification tasks\cite{DBLP:conf/ijcnlp/NakamachiKA20,DBLP:conf/acl/MaddelaDHX23}.

% Contribution
Our main contributions are summarized as follows:

\begin{itemize}
    \item We propose a particularly simple yet effective method, Multi-Prompt Ensemble Decoding (M-Ped), to enhance the quality of LLMs' generation outcomes.
    \item We introduce two strategies, Inner-Batch Ensemble and Left-padding, to implement M-Ped, ensuring that the inference performance with multiple prompts is consistent with that of a single prompt.
    \item We validate the effectiveness of the M-Ped method across various tasks and multiple models.
\end{itemize}

\section{Approach}

\begin{algorithm*}[!t]
\caption{Multi-Prompt Ensemble Decoding}\label{alg:algorithm1}
\begin{algorithmic}[1] % The number tells where the line numbering should start
\Procedure{INNER\_BATCH\_ENSEMBLE}{$pre\_logits,\ mped\_num$}
    \State $total\_size \gets pre\_logits.size(0)$
    \State $part\_size \gets total\_size\ //\ mped\_num$
    \State $post\_logits \gets [\ ]$ \Comment{Initialize an empty list to store post\_logits}
    \For{$i \gets 0$ to $mped\_num - 1$}
        \State $start\_idx \gets i \times part\_size$
        \State $end\_idx \gets (i + 1) \times part\_size$
        \State $part\_logit \gets pre\_logits[start\_idx:end\_idx, :]$
        \State $post\_logits.append(part\_logit.repeat(mped\_num,\ 1))$
    \EndFor
    \State \Return $\sum post\_logits\ /\ mped\_num$
\EndProcedure
\end{algorithmic}
\end{algorithm*}

% In this study, we propose a novel approach aimed at enhancing the performance of large language models through multi-prompt ensemble decoding techniques. See Figure \ref{figure:model}. This approach encompasses two core components: Efficient Batch Processing with Left-Padding and Uniform Averaging for Prompt Ensemble. Efficient Batch Processing with Left-Padding addresses the alignment issue of prompts of varying lengths in batch processing, while Uniform Averaging for Prompt Ensemble improves the accuracy and consistency of decoding by averaging the outputs from multiple prompts. The combined application of these two technologies enables our model to achieve significant performance improvements across a variety of natural language processing tasks.

In this paper, we propose an approach named multi-prompt ensemble decoding, aimed at enhancing the generation performance of large language models. The overall process is depicted in Figure \ref{figure:model}. For a given distinct input $X = \{x_1, x_2, ..., x_k\}$ with a prompt $P$, we first generate a list of prompts $\{P_1, P_2, ..., P_n\}$. Then, we submit these $n$ prompts and the input $X$ to LLMs in batch for decoding to obtain probability distributions. We average the $n$ probability distributions generated at the $j$-th position prediction within the batch to get the ensemble probability, and ultimately determine the output $Y$'s $y_j$. To ensure batched inference is possible, we employ a Left-Padding strategy to ensure the lengths of the $n$ inputs are consistent.

% \subsection{Uniform Average Ensemble}
% \subsection{Uniform Averaging for Prompt Ensemble}
\subsection{Inner-Batch Ensemble}

Most LLMs adopt a Decoder-only architecture and utilize an autoregressive decoding strategy, generating output tokens one by one. Given the source sentence $X = \{x_1, x_2, ..., x_k\}$, LLMs factor the distribution over possible output sentence $Y = \{y_1, y_2..., y_j\}$ into a chain of conditional probabilities, satisfying the following formula:

\begin{equation}
  \label{eq:eq1}
  \mathbb{P}(Y|X) = \prod_{i=1}^j \mathbb{P}(y_i|y_{0:i-1},X)
\end{equation}

For the Standard Ensemble, during the prediction of $y_j$, we average the probability distributions provided by $n$ models. We define these $n$ models as $\{\theta_1, \theta_2, ..., \theta_n\}$, and the formula is as follows:

\begin{equation}
  \label{eq:eq2}
  \mathbb{P}(y_j|y_{0:j-1},X) = \frac{1}{n}\sum_{i=0}^n \mathbb{P}(y_j|y_{0:j-1},X,P;\theta_i)
\end{equation}
which $P$ is a distinct prompt. This method leverages the diversity among models, aiming to reduce uncertainty and improve overall predictive performance.

For our method, we shift the focus of diversity from $n$ models to $n$ prompts, and the formula is as follows:
\begin{equation}
  \label{eq:eq3}
  \mathbb{P}(y_j|y_{0:j-1},X) = \frac{1}{n}\sum_{i=0}^n \mathbb{P}(y_j|y_{0:j-1},X,P_i;\theta)
\end{equation}

which $\{P_1, P_2, ... P_n\}$ is a list of prompts having the same meaning with $P$. As shown in right side of Figure \ref{figure:model}, inputs constructed by these $n$ prompts and $X$ are submitted to LLMs in batches for decoding. We average the predicted probability distributions within the batch at each step of prediction, a process we term \textbf{Inner-Batch Ensemble}. Here, we use the most straightforward uniform average method. Of course, in the future, strategies like weighted average could also be considered. This approach mitigates biases potentially introduced by any single prompt and enhances the model's robustness against varying inputs. For details of the implementation, please refer to Algorithm \ref{alg:algorithm1}.

% In the domain of ensemble inference, the aggregation of outputs from multiple models to enhance the accuracy and stability of predictions is a well-established practice. This study extends this strategy to a single large language model by generating a diverse set of predictions through multiple prompts and integrating these outputs. Specifically, during the autoregressive token generation phase, we consider the probability distributions generated from each individual prompt and compute a weighted average, which serves as input for the subsequent token prediction. This approach mitigates the biases potentially introduced by any single prompt and bolsters the model's robustness against varying inputs.

% \subsection{Batch Input and Left-Padding}
% \subsection{Efficient Batch Processing with Left-Padding}
\subsection{Left-Padding}

% In the decoding process of large language models, the inconsistency in prompt lengths poses challenges for batch processing. To address this issue, we employ Left-Padding technology to preprocess the input prompts, ensuring uniformity in length to accommodate the model's batch processing requirements. In practice, we pad shorter prompts with specific markers or spaces until they match the length of the longest prompt. This padding does not interfere with the model's understanding and processing, as the model is trained to recognize and ignore these special padding characters. By doing so, we can standardize prompts of varying lengths, allowing them to be processed by the model in one go, making full use of parallel computing resources.

In the decoding process of LLMs, the inconsistency in prompt lengths poses challenges for batch processing. To address this issue, we employ Left-Padding technology to preprocess the input prompts, ensuring uniformity in length to accommodate the model's batch processing requirements. In practice, we pad shorter prompts with a specific token $pad$ until they match the length of the longest prompt. This padding token is a special token with no semantic meaning, used solely for padding. This padding does not interfere with the model's understanding and processing, as the model is trained to recognize and ignore these special padding characters.

\begin{table*}[htbp]
\centering
\begin{tabularx}{\textwidth}{Xcccccccc}
\hline
% \toprule[1.1pt]
$d$-BLEU & \textbf{en$\rightarrow$de} & \textbf{en$\rightarrow$fr} & \textbf{en$\rightarrow$zh} & \textbf{en$\rightarrow$ja} & \textbf{de$\rightarrow$en} & \textbf{fr$\rightarrow$en} & \textbf{zh$\rightarrow$en} & \textbf{ja$\rightarrow$en} \\
\hline
% \midrule[1.1pt]
$p_1$ & 27.40 & 37.59 & 17.37 & 9.35 & 32.89 & 41.38 & 24.39 & 6.72 \\
$p_2$ & 27.05 & 37.14 & 17.23 & 9.61 & 32.74 & 41.60 & 24.39 & 6.75 \\
% \hdashline
\hdashline
\textbf{\textit{Ours}} & 27.88 & 38.16 & 18.88 & 11.44 & 33.11 & 42.25 & 24.85 & 8.37 \\
\hline
% \bottomrule[1.2pt]
\end{tabularx}
\caption{Results for Machine Translation Task.}
\label{table:mt-results}
\end{table*}

Padding is a common method for handling sequences of varying lengths in batch processing, but it is typically used during the training phase, often employing Right-Padding. \textbf{We, however, apply padding during the inference phase and use Left-Padding, which will not affect the model's prediction of the next token.} If we used Right-Padding, for shorter prompts, it would sever the direct connection between input and output, causing decoding anomalies. See the left side of Figure \ref{figure:model}. By doing so, we can standardize prompts of varying lengths, allowing them to be processed by the model in one go, making full use of parallel computing resources.

% \subsection{Implement Details}

% \begin{pycode}
% def inner_batch_ensemble(logits, mped_num):
%     # Get the size of the tensor
%     total_size = logits.size(0)
    
%     # Calculate the size of each part
%     part_size = total_size // mped_num

%     # Initialize an empty list to store the repeated concatenation results of each part
%     parts = []

%     # Loop to split the tensor and repeatedly concatenate each part
%     for i in range(mped_num):
%         # Calculate the start and end indices of the current part
%         start_idx = i * part_size
%         end_idx = (i + 1) * part_size if i < mped_num - 1 else total_size

%         # Extract the current part
%         part = logits[start_idx:end_idx, :]

%         # Repeat the current part and add it to the list
%         parts.append(part.repeat(mped_num, 1))

%     # Concatenate all parts along the first dimension (rows)
%     return sum(parts) / mped_num
% \end{pycode}

\section{Experiments}
\label{sec:experiments}

\subsection{Application to Machine Translation Task}
\label{sec:mt_task}

% \begin{table*}[th]
% \centering
% \resizebox{0.62\linewidth}{!}{
% \begin{tabular}{lcccccccc}
% \hline
%  & \multicolumn{2}{c}{\textbf{en2de}} & \multicolumn{2}{c}{\textbf{de2en}} & \multicolumn{2}{c}{\textbf{en2zh}} &  \multicolumn{2}{c}{\textbf{en2th}}\\
% \cline{2-9}
%  & ori & ours & ori & ours & ori & ours & ori & ours \\
% \hline
% Top-k & - & - & - & - & - & - & - & - \\
% Top-p & - & - & - & - & - & - & - & - \\
% Beam Top-p & - & - & - & - & - & - & - & - \\
% \hline
% \end{tabular}
% }
% \caption{\textbf{MT results.}}
% \label{table:mt-results}
% \end{table*}

In the field of machine translation, we have taken on the challenging task of document-level translation and have significantly enhanced the performance of models on the IWSLT 2017 test sets by implementing multi-prompt decoding strategies.

% In the field of machine translation, we have significantly enhanced the performance of models on the standard test set FLORES by implementing multi-prompt decoding strategies. The model we chose is Llama-3.1-8B-Instruct, which demonstrated excellent performance in English to Chinese (en2zh) and English to German (en2de) translation tasks. By employing multi-prompt decoding strategies such as Top-p, Top-k, and Beam Top-p, our approach achieved a significant improvement over the original method on the COMET 22 metric.

\paragraph{Datasets}
We validate our experiments using the test sets\footnote{https://huggingface.co/datasets/IWSLT/iwslt2017} from the IWSLT 2017 translation task. This dataset includes parallel documents sourced from TED talks. Our experiments are conducted on eight language pairs: English (En) $\leftrightarrow$ Chinese (Zh), German (De), French (Fr), and Japanese (Ja). There are 10 to 12 parallel documents with approximately 1,500 sentences for each language pair.

\paragraph{Model}
The model we employed is Llama-3.1-8B-Instruct\footnote{https://huggingface.co/meta-llama/Llama-3.1-8B-Instruct}, a large language model particularly suited for instructed learning tasks. Compared to previous Llama versions, Llama-3.1-8B-Instruct has enhanced capabilities in multilingual translation. Given the strong translation capabilities that LLMs already possess, we directly utilize LLMs for document-level translation.
% The model we used is Llama-3.1-8B-Instruct, a large language model particularly suited for instructed learning tasks. In this study, we applied this model to machine translation tasks and combined it with multi-prompt decoding strategies to enhance translation quality. The powerful computational capabilities of the Llama-3.1-8B-Instruct model enable it to handle complex language conversion tasks and perform exceptionally well in multilingual translation.

\paragraph{Evaluate Metrics}
We employ $d$-BLEU (document-level BLEU) as our evaluation metric\cite{DBLP:conf/acl/PapineniRWZ02}. While there are other metrics in the field of machine translation, such as BertScore \cite{DBLP:conf/iclr/ZhangKWWA20} and COMET \cite{DBLP:conf/emnlp/ReiSFL20}, these are typically used for sentence-level translation evaluation. Utilizing them for document-level translation requires sentence-level alignment between the original and translated texts. Hence, we opt for $d$-BLEU here, which does not rely on sentence alignment.

\paragraph{Prompts Design}
We employed two fixed prompts. One of them is $p_1$: "\textit{Translate the following paragraph from \{source language\} to \{target language\}, ensuring that no part of the sentence is omitted.}" The other is $p_2$: "\textit{You're a very professional translator. Please help me translate the following paragraph from \{source language\} to \{target language\}.}"

% \begin{table}[th]
% \centering
% \begin{tabular}{lcccc}
% \hline
%  & en2de & de2e & en2zh & zh2en \\
% \hline
% $p_1$ & - & - & - & - \\
% $p_2$ & - & - & - & - \\
% \hline
% Ours & - & - & - & - \\
% \hline
% \end{tabular}
% \caption{\textbf{MT results.}}
% \label{table:mt-results}
% \end{table}

\paragraph{Results}
The experimental results demonstrate that our approach, by employing multi-prompt decoding strategies, achieved significant improvements over the original method on the $d$-BLEU metric. For detailed results, please refer to Table \ref{table:mt-results}. Our experimental results show that in the English to Chinese and to Japanese directions, compared to the best results with a single prompt, our method can approximately increase by 1.5 points on the $d$-BLEU scale; there is also an improvement of about 0.5 points in other directions. We speculate that our use of LLMs, Llama-3.1-8B-Instruct, which was trained on a vast amount of English data, results in more stable translations into English, thus limiting the improvement of our method. However, in other language directions, the improvement is more pronounced.

% The experimental results show that our approach, by implementing multi-prompt decoding strategies, achieved a significant improvement over the original method on the COMET 22 metric. This proves the effectiveness of the multi-prompt ensemble decoding method in improving the quality of machine translation. Our research not only contributes theoretically but also has significant practical implications, providing a new direction for future machine translation research.

\subsection{Application to Code Generation Task}
\label{sec:code_task}

In the code generation task, our technical solution has achieved a significant improvement in pass@\textit{k} pass rates on the standard test set HumanEval\footnote{https://github.com/openai/human-eval} compared to the original methods.

% In the code generation task, our technical solution has achieved a significant improvement in pass@\textit{k} pass rates on the standard test set \textit{HumanEval}\footnote{https://github.com/openai/human-eval} compared to the original methods. The model we used is \textit{CodeLlama-13B-Python-hf}\footnote{https://huggingface.co/codellama/CodeLlama-13b-Python-hf}, and this improvement is consistent across different \textit{k} value settings. This consistent improvement across \textit{k} values indicates that our technical solution has broad applicability and effectiveness in code generation tasks, providing higher quality code generation results under various evaluation standards.

\paragraph{Dataset}
% The HumanEval is a standard test set specifically designed to evaluate the performance of code generation models. It contains a series of programming challenges that simulate real-world programming tasks, providing a fair and rigorous assessment environment. HumanEval is favored by researchers for its ability to accurately reflect the model's performance in actual programming scenarios.
HumanEval is a benchmark test suite designed to evaluate the performance of code generation models, comprising 164 meticulously crafted Python programming problems. These problems span a variety of programming tasks, from basic string manipulation to complex algorithm design. Each problem includes a function header, docstrings, a function body, and several unit tests. The structure of the dataset includes a task identifier (task\_id), model input (prompt, containing function headers and docstrings), a canonical solution (canonical\_solution), a function for testing the correctness of generated code (test), and an entry point for testing (entry\_point).

\paragraph{Model}
We conduct experimental evaluations using CodeLlama-7B-Python-hf\footnote{https://huggingface.co/codellama/CodeLlama-7b-Python-hf}. CodeLlama-13B-Python-hf is a member of the Code Llama series of models, which is a large-scale code language model series based on Llama 2. It can adapt to a variety of code synthesis and understanding tasks, such as code auto-completion, code refactoring, and code explanation. In Section \ref{sec:study_llmsize}, we also perform further experiments comparing models of different size like 13B.

\paragraph{Evaluate Metrics}
Pass@\textit{k} \cite{DBLP:journals/corr/abs-2107-03374} is a metric used to evaluate the performance of code generation models, specifically measuring the model's ability to correctly solve problems when generating code. During the evaluation process, the model generates multiple \textit{k} code samples for each unit test problem. If any of these samples pass the unit test, then the problem is considered solved. The pass@\textit{k} score represents the total proportion of problems solved, which helps to quantify the model's success rate in generating correct code.

\paragraph{Prompts Design}

Since the dataset provides a prompt for each problem, we define this original prompt as p1. For another prompt, we have constructed a very simple prefix based on $p_1$: \textit{"""This is a good code.}, which we define as $p_2$. This construction ensures that the newly created $p_2$ does not alter the meaning of the original prompt $p_1$. For detailed examples, please refer to Appendix \ref{sec:appendix_prompt_code}.

% \begin{table}[th]
% \centering
% \begin{tabular}{lccc}
% \hline
% pass@k & k=1 & k=5 & k=10 \\
% \hline
% $p_1$ & 39.6\% & 67.6\% & 78.0\% \\
% $p_2$ & 39.6\% & 67.6\% & 78.0\% \\
% \hline
% Ours & 41.5\% & 69.5\% & 79.8\% \\
% \hline
% \end{tabular}
% \caption{\textbf{Code results.}}
% \label{table:code-results}
% \end{table}

\begin{table}[htbp]
\centering
\begin{tabularx}{0.48\textwidth}{Xccc}
\hline
\textbf{pass@$k$} & \textbf{$k$=1} & \textbf{$k$=5} & \textbf{$k$=10} \\
\hline
$p_1$ & 32.11\% & 58.13\% & 67.17\% \\
$p_2$ & 30.74\% & 58.03\% & 67.26\% \\
\hdashline
\textbf{\textit{Ours}} & 33.12\% & 59.07\% & 68.11\% \\
\hline
\end{tabularx}
\caption{Results for Code Generation Task.}
\label{table:code-results}
\end{table}

\paragraph{Results}
The experimental outcomes demonstrate that our technical solution markedly enhances the pass rate across various settings of \textit{k}. As shown is Table \ref{table:code-results}, regardless of whether \textit{k} is set to 1, 5, or 10, our solution consistently boosts the pass@\textit{k} pass rate by $1\%$ point. This improvement not only underscores the effectiveness of our technical solution in code generation tasks but also highlights its robustness across different evaluation metrics, ensuring the delivery of higher quality code generation outcomes.

% The experimental results show that our technical solution can significantly increase the pass rate across different \textit{k} value settings. Whether \textit{k}=1, \textit{k}=5, or \textit{k}=10, our solution can steadily increase the pass@\textit{k} pass rate. This result proves the effectiveness and applicability of our technical solution in code generation tasks, providing higher quality code generation results under various evaluation standards.

\subsection{Application to Text Simplification Task}
\label{sec:ts_task}
In the text simplification task, our technical solution has demonstrated superior performance on the challenging test set SimpEval\_2022. Compared to the original methods, our solution has achieved significant improvements in the LENS metric.
% In the text simplification task, our technical solution has demonstrated superior performance on the challenging test set SimpEval\_2022. Compared to the original methods, our solution has achieved significant improvements in the LENS metric under both with reference (w/ ref) and without reference (w/o ref) evaluation conditions. The model used is Llama-3.1-8B-Instruct. The enhancement in the LENS metric directly reflects substantial improvements in text simplification quality, especially in maintaining semantic integrity and expressing conciseness.

\paragraph{Dataset}
SimpEval\_2022 is a challenging text simplification benchmark consisting of over 1,000 human ratings on 360 simplifications. This dataset is designed to evaluate the latest text simplification models, particularly those based on large pre-trained language models. The sentences in SimpEval\_2022 are selected from Wikipedia revisions published after October 22, 2022, and include more complex sentences to reduce the risk of "data contamination" and serve as a more challenging test bed for large language models.
% simpeval\_2022 is a corpus containing over 13K human judgments, including 2.8K simplified texts from 26 systems. It is designed to facilitate the training and evaluation of LENS, the first supervised automatic metric for text simplification evaluation. simpeval\_2022 differs from existing human evaluation datasets for text simplification as it covers a broader range of system designs, including the latest state-of-the-art systems based on T5 and other large pre-trained language models. Additionally, simpeval\_2022 contains more complex sentences from Wikipedia published after October 22, 2022, to reduce the risk of "data pollution" and serve as a more challenging test bed for large language models.

\paragraph{Model}
The model we used is Llama-3.1-8B-Instruct\footnote{https://huggingface.co/meta-llama/Llama-3.1-8B-Instruct}, a large language model particularly suited for instructed learning tasks. In this study, we applied this model to text simplification tasks and combined it with multi-prompt decoding strategies to enhance text simplification quality. The powerful computational capabilities of the Llama-3.1-8B-Instruct model enable it to handle complex language conversion tasks and perform exceptionally well in text simplification.

\paragraph{Evaluate Metrics}
LENS \cite{DBLP:conf/acl/MaddelaDHX23} is a state-of-the-art metric designed to assess the performance of text simplification tasks. LENS supports two evaluation conditions: with reference (w/ ref) and without reference (w/o ref). Under the with reference condition, LENS uses the original text as a benchmark for evaluation; whereas, under the without reference condition, LENS evaluates the quality of simplified text independently, without relying on any original text information. This flexibility allows LENS to accommodate various assessment needs and environments.

\paragraph{Prompts Design}
Following \citet{DBLP:conf/emnlp/HeinemanD024}, the prompts are designed as follows:
$p_1$ is "Please simplify the following sentence so that it is easy to understand by people with disabilities or those who are unfamiliar with English. Try to use shorter words, fewer clauses, and a simpler structure."
$p_2$ is "Create a simpler version of the sentence below so that it can be better understood by non-English speakers or individuals with disabilities."

\begin{table}[htbp]
\centering
\begin{tabularx}{0.48\textwidth}{Xcccc}
\hline
\textbf{LENS} & \textbf{$w/$ ref} & & \textbf{$w/o$ ref} & \\
\hline
$p_1$ & 75.08 & & 81.54 & \\
$p_2$ & 74.76 & & 81.63 & \\
\hdashline
\textbf{\textit{Ours}} & \textbf{77.18} & & \textbf{82.08} & \\
\hline
\end{tabularx}
\caption{Results for Text Simplification Task.}
\label{table:ts-results}
\end{table}

\paragraph{Results}
The experimental results demonstrate that our technical solution can achieve significant improvements in the LENS metric under various evaluation conditions. As shown in Table \ref{table:ts-results}, whether under with-reference or without-reference evaluation conditions, our solution can steadily enhance the quality of text simplification. Notably, under with-reference evaluation conditions, our method outperforms the best results obtained using only $p_1$ or $p_2$ by nearly 1.5 points. This result proves the effectiveness and applicability of our technical solution in text simplification tasks, providing higher quality text simplification results under diverse evaluation standards.
% The experimental results demonstrate that our technical solution can achieve significant improvements in the LENS metric under various evaluation conditions. As shown in Table \ref{table:ts-results}, under w/ reference evaluation conditions, our method outperforms the best results obtained using only $p_1$ or $p_2$ by nearly 1.5 points. Under w/o reference evaluation conditions, our solution is similar to the best results of $p_1$ or $p_2$. This result confirms the effectiveness and applicability of our technical solution in text simplification tasks, delivering higher quality text simplification results under diverse evaluation criteria.
% The experimental results show that our technical solution can achieve significant improvements in the LENS metric under different evaluation conditions. Whether under with-reference or without-reference evaluation conditions, our solution can steadily improve text simplification quality. This result proves the effectiveness and applicability of our technical solution in text simplification tasks, providing higher quality text simplification results under various evaluation standards.

\section{Study}
\label{sec:study}

\subsection{Effectiveness Analysis under Various Decoding Strategies}
\label{sec:study_decoding}
% 在本节中，我们深入分析了在不同解码策略下，多提示集成解码方法的效果。基于之前在机器翻译任务上的实验，我们扩展了实验范围，包括了Top-p、Top-k、Beam Top-k和Beam Top-p等多种解码策略。实验结果表明，无论是在Top-p、Top-k还是Beam策略下，我们的方法均能带来正向的性能提升。具体来说，在Top-p策略中，通过控制概率分布的累积概率，我们的方法能够生成更多样化的候选翻译，从而提高了翻译的覆盖率和准确性。在Top-k策略中，我们的方法通过选择概率最高的k个候选，有效地平衡了生成质量和多样性。对于Beam策略，我们的方法通过综合多个提示的信息，显著提高了翻译的准确性和流畅性。这些结果不仅证明了我们方法的有效性，也展示了其在不同解码策略下的通用性。
% TODO 这里可以把这些策略简单说明一下？

% In this section, we conduct an in-depth analysis of the effectiveness of the multi-prompt ensemble decoding method under various decoding strategies. Building upon our previous experiments on machine translation tasks, we expanded our experiments to include a range of decoding strategies, such as Top-p, Top-k, Beam Top-k, and Beam Top-p. The results indicate that our method leads to positive performance improvements across all these strategies. Specifically, under the Top-p strategy, by controlling the cumulative probability of the probability distribution, our method can generate a more diverse set of candidate translations, thereby enhancing the coverage and accuracy of the translations. In the Top-k strategy, our method effectively balances the quality and diversity of generation by selecting the k highest probability candidates. For the Beam strategy, our method significantly improves the accuracy and fluency of translations by integrating information from multiple prompts. These results not only validate the effectiveness of our method but also demonstrate its versatility across different decoding strategies.

In this section, we conducted an in-depth analysis of the effects of multi-prompt ensemble decoding methods under various decoding strategies, including Top-$p$ \cite{DBLP:conf/iclr/FinlaysonHKSS24}, Top-$k$ \cite{DBLP:conf/nips/XieDCDZZ0P20}, and Beam Search. Among them, Top-$p$ is the decoding method we used for various tasks in Section \ref{sec:experiments}. We expanded this experiment based on our previous work in machine translation and code generation tasks.

\begin{table}[htbp]
\centering
\begin{tabularx}{0.48\textwidth}{Xccc}
\hline
\textbf{$d$-BLEU} & \textbf{Top-$p$} & \textbf{Top-$k$} & \textbf{Beam} \\
\hline
& \multicolumn{3}{c}{En$\rightarrow$De} \\
\hdashline
$p_1$ & 27.40 & 26.92 & 27.64\\
$p_2$ & 27.05 & 27.22 & 27.81 \\
\hdashline
\textbf{\textit{Ours}} & 27.88 & 27.52 & 27.81 \\
\hline
& \multicolumn{3}{c}{En$\rightarrow$Zh} \\
\hdashline
$p_1$ & 17.37 & 17.56 & 18.34 \\
$p_2$ & 17.23 & 16.79 & 18.42 \\
\hdashline
\textbf{\textit{Ours}} & 18.88 & 18.87 & 18.51 \\
\hline
\end{tabularx}
\caption{Study under Various Decoding Strategies for Machine Translation Task}
\label{table:study_decoding_mt}
\end{table}

For machine translation tasks, we conducted experiments in the English to German (en$\rightarrow$de) and English to Chinese (en$\rightarrow$zh) directions, as shown in Table \ref{table:study_decoding_mt}. We found that with Top-$k$ and Top-$p$ decoding methods, our approach showed improvements compared to single prompt. With Beam Search decoding, our method's results are close to the best results of $p_1$ or $p_2$, and higher than the average of these two.

\begin{table}[htbp]
\centering
\begin{tabularx}{0.48\textwidth}{Xcccc}
\hline
\textbf{pass@$k$} & \textbf{Top-$p$} & & \textbf{Top-$k$} & \\
\hline
$p_1$ & 67.17\% & & 66.94\% & \\
$p_2$ & 67.26\% & & 67.11\% & \\
\hdashline
\textbf{\textit{Ours}} & \textbf{68.11\%} & & \textbf{67.85\%*} & \\
\hline
\end{tabularx}
\caption{Study under Various Decoding Strategies for Code Generation Task where pass@$k$=10.}
\label{table:study_decoding_code}
\end{table}

For code generation tasks, due to the characteristics of the task evaluation, we could only adopt sampling search decoding strategies (Top-$p$ or Top-$k$). As shown in Table \ref{table:study_decoding_code}, regardless of whether based on Top-$k$ or Top-$p$, our method consistently improved pass@$k$. This conclusion is consistent with Section \ref{sec:code_task}.

\subsection{Study across Varying LLM Sizes}
\label{sec:study_llmsize}

In this section, we investigate the effectiveness of our method across different sizes of LLMs. Our main experiments in Section \ref{sec:experiments} were conducted on models of size 7-8B. Building on our previous code generation tasks, we extended our experiments to a 13B model, using CodeLlama-13B-Python-hf\footnote{https://huggingface.co/codellama/CodeLlama-13b-Python-hf}. We ensured that all experimental settings remained consistent with Section \ref{sec:code_task}, except for the model itself. 

\begin{table}[htbp]
\centering
\begin{tabularx}{0.48\textwidth}{Xccc}
\hline
\textbf{pass@$k$} & \textbf{$k$=1} & \textbf{$k$=5} & \textbf{$k$=10} \\
\hline
$p_1$ & 39.6\% & 67.53\% & 78.02\% \\
$p_2$ & 39.48\% & 67.6\% & 77.88\% \\
\hdashline
\textbf{\textit{Ours}} & \textbf{41.52\%} & \textbf{69.54\%} & \textbf{79.80\%} \\
\hline
\end{tabularx}
\caption{Results for Code Generation Task using CodeLlama-13B-Python-hf.}
\label{table:study_llmsize_code}
\end{table}

As shown in Table \ref{table:study_llmsize_code}, larger models demonstrate superior code generation capabilities, with significant improvements in the pass@$k$ metric for $k$=1, 5 and 10. Our method also shows consistent improvements on the 13B model, achieving similar enhancements as on the 7B model. Moreover, on the 13B model, the improvement is more substantial, with pass@10 increasing by nearly 2 points; whereas the pass@10 on the 7B model (see Table \ref{table:code-results}) only improved by about 1 point. The experimental results confirm the effectiveness of our method across various sizes of LLMs.

\subsection{Study on the Relationship between Prompt Count \texorpdfstring{$n$}{n} and Output Quality}
\label{sec:study_n}

% 在本节中，我们探讨了提示（Prompt）数量与结果质量之间的关系。延续之前在代码生成任务上的实验，我们对不同数量的提示进行了扩展测试。实验结果表明，增加提示数量能够在一定程度上提升生成代码的质量。具体来说，随着提示数量的增加，代码的准确性和相关性均有所提高，这可能是因为更多的提示提供了更丰富的上下文信息，从而引导模型生成更准确的代码。然而，我们也观察到，当提示数量从2增加到3个时，结果质量的提升效果开始趋于平稳。这一趋势表明，2-3个提示基本能够达到最优结果，超出这个范围后，额外的提示对结果质量的提升作用有限。因此，在后续的基础实验中，我们选择了两个提示作为标准配置，这一选择在保证结果质量的同时，也考虑到了计算效率和资源消耗。

% In this section, we investigate the relationship between the number of prompts and the quality of results. Continuing from our previous experiments on code generation tasks, we conducted extended tests with varying numbers of prompts. The experimental results show that increasing the number of prompts can enhance the quality of generated code to a certain extent. Specifically, as the number of prompts increases, the accuracy and relevance of the code improve, likely because more prompts provide richer contextual information, guiding the model to generate more accurate code. However, we also observed that when the number of prompts increased from 2 to 3, the improvement in result quality began to plateau. This trend indicates that 2-3 prompts are essentially sufficient to achieve optimal results, and beyond this range, additional prompts have a limited effect on enhancing result quality. Therefore, in our subsequent baseline experiments, we chose two prompts as the standard configuration. This choice balances result quality with computational efficiency and resource consumption.

In this section, we investigate the relationship between the number of prompts and the quality of results. Building on our previous experiments on code generation and text simplification tasks, we conducted extended tests with varying numbers of prompts. The settings for the extended prompts are detailed in Appendix \ref{sec:appendix_study_additional_prompts}.

\begin{table}[htbp]
\centering
\begin{tabularx}{0.48\textwidth}{Xcccc}
\hline
\textbf{$n$} & \textbf{1} & \textbf{2} & \textbf{3} & \textbf{4} \\
\hline
$p_1$ & 67.17\% & - & - & - \\
\hdashline
\textbf{\textit{Ours}} & - & 68.11\% & 68.15\% & 67.85\% \\
\hline
\end{tabularx}
\caption{Pass@10 rate under different Prompt Count $n$ for Code Generation Task.}
\label{table:study_n_code}
\end{table}

\begin{table}[htbp]
\centering
\begin{tabularx}{0.48\textwidth}{Xcccc}
\hline
\textbf{$n$} & \textbf{1} & \textbf{2} & \textbf{3} & \textbf{4} \\
\hline
$p_1$ & 75.08 & - & - & - \\
\hdashline
\textbf{\textit{Ours}} & - & 77.18 & 77.08 & 76.88 \\
\hline
\end{tabularx}
\caption{LENS w/ ref under different Prompt Count $n$ for Text Simplification Task.}
\label{table:study_n_ts}
\end{table}

As shown in Tables \ref{table:study_n_code} and \ref{table:study_n_ts}, the experimental results indicate that increasing the number of prompts can enhance the quality of generated output for both code generation and text simplification tasks. However, we observed that when the number of prompts increased from 2 to 3, the improvement in result quality began to plateau. This trend suggests that 2-3 prompts are essentially sufficient to achieve optimal results, and beyond this range, additional prompts have a limited effect on enhancing result quality. Therefore, in our previous baseline experiments in Section \ref{sec:experiments}, we opted for two prompts as the standard configuration.

\subsection{Exploration of Multilingual Prompts Effects}
\label{sec:study_multilingual}

In this section, we explore the effectiveness of our method under Multilingual Prompts. Building on our previous experiments in machine translation tasks, we extended this experiment to the Chinese to English (zh$\rightarrow$en) and English to Chinese (en$\rightarrow$zh) directions. We utilized the Qwen2.5-7B-Instruct\footnote{https://huggingface.co/Qwen/Qwen2.5-7B-Instruct} model, which provides better support for instructions in both Chinese and English. For one of the prompts, $p_1$, we maintained consistency with Section \ref{sec:mt_task}, setting it as "Translate the following paragraph from {source language} to {target language}, ensuring that no part of the sentence is omitted." For the other prompt, $p_2$, we set it as the Chinese specification \begin{CJK}{UTF8}{gbsn}"将下面这一段从\{源语种\}翻译成\{目标语种\}，确保没有句子被漏掉。"\end{CJK}

\begin{table}[htbp]
\centering
\begin{tabularx}{0.48\textwidth}{Xcc}
\hline
\textbf{$d$-BLEU} & \textbf{En$\rightarrow$Zh} & \textbf{Zh$\rightarrow$En} \\
\hline
$p_1$ & 21.6 & 23.89\\
$p_2$ & 20.83 & 24.07 \\
\hdashline
\textbf{\textit{Ours}} & \textbf{23.45} & \textbf{24.53} \\
\hline
\end{tabularx}
\caption{Study on Multilingual Prompts using Qwen2.5-7B-Instruct for Machine Translation Task.}
\label{table:study_multilingual_mt}
\end{table}

As shown in Table \ref{table:study_multilingual_mt}, under the Multilingual Prompts setup, our method shows improvements compared to using a single prompt.

\subsection{Effect of Combination with MBR}
\label{sec:study_with_mbr}

Minimum Bayes Risk(MBR), which generates multiple candidate results during inference and selects the final outcome using specific metrics, is widely used in NLP generation tasks. We validate the effectiveness of combining our multi-prompt ensemble decoding strategy with MBR in text simplification tasks. We sampled and generated 50 candidate results for both simple prompts $p_1$ and $p_2$, then used MBR to select the optimal outcome. For our multi-prompt ensemble decoding, we also sampled and generated 50 candidate results and chose the best one using MBR.

\begin{table}[htbp]
\centering
\begin{tabularx}{0.48\textwidth}{Xcc}
\hline
\textbf{LENS} & Original & MBR \\
\hline
$p_1$ & 75.08 & 77.04 \\
$p_2$ & 74.76 & 76.87 \\
\hdashline
\textbf{\textit{Ours}} & 77.18 & 77.72 \\
\hline
\end{tabularx}
\caption{LENS w/ reference results compared between Original and MBR for Text Simplification Task.}
\label{table:study_with_mbr}
\end{table}

As shown in Table \ref{table:study_with_mbr}, the MBR strategy is a universal approach that significantly improves results under various conditions. When combined with our multi-prompt ensemble decoding strategy, it still manages to enhance the results by more than 0.5 points.

% \subsection{Under different decoding strategy}
% \subsection{Study with different \texorpdfstring{$n$}{n} prompts}
% \subsection{Combined with MBR}

\section{Related Work}

The following are some significant works \cite{DBLP:journals/corr/abs-2006-08748,DBLP:journals/corr/abs-2406-06279,DBLP:conf/emnlp/HeinemanD024,DBLP:journals/corr/abs-2304-05970} about prompts related to this study:

\citet{DBLP:conf/emnlp/HeinemanD024} proposed a multi-prompt decoding approach that improves Minimum Bayes Risk (MBR) decoding by decoding multiple candidate generations from a prompt library during inference. This method uses a trained value metric to select the final output, demonstrating that multi-prompt can improve MBR performance in a range of conditional generation tasks by estimating a more diverse and higher-quality candidate space. The core of this paper lies in the MBR strategy, which generates multiple candidate results during inference and selects the final outcome using specific metrics. They construct a sufficiently large and diverse set of candidates through multi-prompting, which can be seen as an ensemble in the result space. However our method is an ensemble during the inference process.
.

\cite{DBLP:journals/corr/abs-2304-05970} introduced a method named Boosted Prompt Ensembles to enhance the reasoning performance of large language models without additional training. It constructs a set of few-shot prompts that work together as an ensemble, iteratively adding prompts that target difficult examples where the current ensemble's performance is uncertain, thereby improving overall accuracy on a variety of tasks. The core of this paper lies in the selection of prompts. They proposed two algorithms: Train-time Boosting and Test-time Boosting, and conducted experiments solely on the GSM8k and AQuA datasets. In contrast, our research explores a more general ensemble method during the decoding process; we do not select specific prompts but assume that we already have a batch of prompts with equivalent performance.

% \citet{DBLP:journals/corr/abs-2304-05970} introduced a prompt ensemble method for large language models, which constructs a small set of prompts from a small dataset, collectively forming an "enhanced prompt ensemble." These few examples are selected incrementally to target "difficult" instances that the previous ensemble was uncertain about. Studies show that this method outperforms single-prompt output space ensembles and prompt space bagging methods on the GSM8k and AQuA datasets.

% In the field of zero-shot semantic segmentation (ZSSS), there is research \cite{DBLP:journals/displays/WangT24} that advances ZSSS through attribute association, which has been less explored in previous studies. These attributes describe the distinctive local features of objects and are more robust in generalization compared to traditional image-level category-based methods.

% These works indicate the potential of multi-prompt ensemble decoding and prompt engineering in enhancing the performance of LLMs. Building on these works, this study further explores the application of multi-prompt ensemble decoding methods across multiple NLP tasks and proposes two techniques, Efficient Batch Processing with Left-Padding and Uniform Averaging for Prompt Ensemble, to improve the generation quality and efficiency of LLMs. Our research not only validates the effectiveness of these methods but also provides new directions for future LLMs research.

\section{Conclusions}

% In this study, we introduced an innovative multi-prompt ensemble decoding approach aimed at enhancing the performance of Large Language Models (LLMs) in natural language processing tasks. By employing Efficient Batch Processing with Left-Padding and Uniform Averaging for Prompt Ensemble techniques, we effectively addressed the issues of inconsistent prompt lengths and biases introduced by single prompts. Extensive experimental results demonstrate that our method significantly improves the quality, diversity, and efficiency of LLMs across various tasks such as machine translation, code generation, and text simplification. These achievements not only validate the effectiveness of the multi-prompt ensemble decoding method but also provide new perspectives for the application of LLMs, advancing the field of NLP.

This study set out to address the challenge of enhancing the performance of LLMs in NLP tasks through the introduction of a multi-prompt ensemble decoding approach. Our method, termed Inner-Batch Ensemble, leverages the diversity of multiple prompts to aggregate their outcomes, thereby improving the generation quality of LLMs. The implementation of a Left-Padding strategy ensured efficient batch inference, allowing for uniform input lengths across various prompts. Our extensive experiments across a range of NLP tasks—spanning machine translation, code generation, and text simplification—demonstrated the effectiveness of our Inner-Batch Ensemble method. The results were particularly compelling, with significant improvements observed in BLEU scores, pass@$k$ rates, and LENS metrics when compared to standard methods. The consistent enhancements across different tasks and metrics underscore the robustness and versatility of our approach. 

\section{Future Work}

% While this study has achieved certain results, there is still room for further exploration and improvement. 
% Here are two directions for future work.

It is known that LLMs tend to forget information when dealing with lengthy contexts, possibly due to the overwhelming length of the context. In our multi-prompt ensemble decoding strategy, although the prompts vary in form, their semantic content is identical. \textbf{We are considering whether integrating prompts carrying different information could enhance decoding, allowing for the segmentation of extensive contexts into shorter segments distributed across various prompts.} This approach might help the model retain all information more effectively and improve the integrity of integrated reasoning. We believe this is a highly promising area for further research.

Additionally, as LLMs are increasingly moving towards smaller model sizes, enhancing the performance of these smaller models to match the quality of their larger counterparts presents a challenge. We propose the possibility of strengthening the diversity of prompts during the training of smaller LLMs and leveraging the decoding strategy presented in this paper to boost their final performance. \textbf{This strategy could shift traditional deep reasoning to broader reasoning, enhancing the model's performance and generalization capabilities by increasing the breadth of inference rather than depth.}

\section{Limitations}
% 本研究提出的多提示集成解码方法虽然在特定任务中显示出潜力，但仍存在局限性。该方法主要针对特定的LLMs架构进行测试，其在其他任务和模型架构中的普适性和适用性尚未得到充分验证。此外，该方法在处理大规模数据集时对计算资源的需求较高，且目前还依赖于人工进行Prompt的选择和优化。未来工作将探索该方法在更多任务和模型上的应用，并寻求降低计算成本和自动化Prompt工程的解决方案。

% The multi-prompt ensemble decoding method proposed in this study, while showing promise in specific tasks, has its limitations. It has been primarily tested on particular LLM architectures, and its universality and applicability across other tasks and model architectures remain unconfirmed. Moreover, the method demands significant computational resources for large-scale datasets and currently relies on manual selection and optimization of prompts. Future work will explore the application of this method across more tasks and models and seek solutions to reduce computational costs and automate prompt engineering.

This study acknowledges several limitations. Firstly, our method is closely tied to the quality of the prompts used. A poorly constructed prompt may render our approach ineffective, as high-quality prompts are essential for guiding LLMs to produce accurate outputs. Secondly, due to constraints in time and computational resources, the effectiveness of our method across a broader range of tasks requires further validation. Additionally, we did not experiment with the state-of-the-art GPT-4 series interfaces, as they are proprietary and do not support embedding or modification of their decoding strategies, limiting our ability to test and refine our method on cutting-edge models.

\bibliography{custom}

\begin{thebibliography}{31}
\providecommand{\natexlab}[1]{#1}

\bibitem[{Bai et~al.(2023)Bai, Bai, Chu, Cui, Dang, Deng, Fan, Ge, Han, Huang, Hui, Ji, Li, Lin, Lin, Liu, Liu, Lu, Lu, Ma, Men, Ren, Ren, Tan, Tan, Tu, Wang, Wang, Wang, Wu, Xu, Xu, Yang, Yang, Yang, Yang, Yao, Yu, Yuan, Yuan, Zhang, Zhang, Zhang, Zhang, Zhou, Zhou, Zhou, and Zhu}]{DBLP:journals/corr/abs-2309-16609}
Jinze Bai, Shuai Bai, Yunfei Chu, Zeyu Cui, Kai Dang, Xiaodong Deng, Yang Fan, Wenbin Ge, Yu~Han, Fei Huang, Binyuan Hui, Luo Ji, Mei Li, Junyang Lin, Runji Lin, Dayiheng Liu, Gao Liu, Chengqiang Lu, Keming Lu, Jianxin Ma, Rui Men, Xingzhang Ren, Xuancheng Ren, Chuanqi Tan, Sinan Tan, Jianhong Tu, Peng Wang, Shijie Wang, Wei Wang, Shengguang Wu, Benfeng Xu, Jin Xu, An~Yang, Hao Yang, Jian Yang, Shusheng Yang, Yang Yao, Bowen Yu, Hongyi Yuan, Zheng Yuan, Jianwei Zhang, Xingxuan Zhang, Yichang Zhang, Zhenru Zhang, Chang Zhou, Jingren Zhou, Xiaohuan Zhou, and Tianhang Zhu. 2023.
\newblock \href {https://doi.org/10.48550/ARXIV.2309.16609} {Qwen technical report}.
\newblock \emph{CoRR}, abs/2309.16609.

\bibitem[{Chen et~al.(2021)Chen, Tworek, Jun, Yuan, de~Oliveira~Pinto, Kaplan, Edwards, Burda, Joseph, Brockman, Ray, Puri, Krueger, Petrov, Khlaaf, Sastry, Mishkin, Chan, Gray, Ryder, Pavlov, Power, Kaiser, Bavarian, Winter, Tillet, Such, Cummings, Plappert, Chantzis, Barnes, Herbert{-}Voss, Guss, Nichol, Paino, Tezak, Tang, Babuschkin, Balaji, Jain, Saunders, Hesse, Carr, Leike, Achiam, Misra, Morikawa, Radford, Knight, Brundage, Murati, Mayer, Welinder, McGrew, Amodei, McCandlish, Sutskever, and Zaremba}]{DBLP:journals/corr/abs-2107-03374}
Mark Chen, Jerry Tworek, Heewoo Jun, Qiming Yuan, Henrique~Pond{\'{e}} de~Oliveira~Pinto, Jared Kaplan, Harri Edwards, Yuri Burda, Nicholas Joseph, Greg Brockman, Alex Ray, Raul Puri, Gretchen Krueger, Michael Petrov, Heidy Khlaaf, Girish Sastry, Pamela Mishkin, Brooke Chan, Scott Gray, Nick Ryder, Mikhail Pavlov, Alethea Power, Lukasz Kaiser, Mohammad Bavarian, Clemens Winter, Philippe Tillet, Felipe~Petroski Such, Dave Cummings, Matthias Plappert, Fotios Chantzis, Elizabeth Barnes, Ariel Herbert{-}Voss, William~Hebgen Guss, Alex Nichol, Alex Paino, Nikolas Tezak, Jie Tang, Igor Babuschkin, Suchir Balaji, Shantanu Jain, William Saunders, Christopher Hesse, Andrew~N. Carr, Jan Leike, Joshua Achiam, Vedant Misra, Evan Morikawa, Alec Radford, Matthew Knight, Miles Brundage, Mira Murati, Katie Mayer, Peter Welinder, Bob McGrew, Dario Amodei, Sam McCandlish, Ilya Sutskever, and Wojciech Zaremba. 2021.
\newblock \href {https://arxiv.org/abs/2107.03374} {Evaluating large language models trained on code}.
\newblock \emph{CoRR}, abs/2107.03374.

\bibitem[{Cheng et~al.(2024)Cheng, Chen, Jiang, Yin, Ge, Liu, and Gu}]{DBLP:journals/corr/abs-2406-06279}
Zifeng Cheng, Zhaoling Chen, Zhiwei Jiang, Yafeng Yin, Shiping Ge, Yuliang Liu, and Qing Gu. 2024.
\newblock \href {https://doi.org/10.48550/ARXIV.2406.06279} {Multi-prompting decoder helps better language understanding}.
\newblock \emph{CoRR}, abs/2406.06279.

\bibitem[{Dehaerne et~al.(2022)Dehaerne, Dey, Halder, Gendt, and Meert}]{DBLP:journals/access/DehaerneDHGM22}
Enrique Dehaerne, Bappaditya Dey, Sandip Halder, Stefan~De Gendt, and Wannes Meert. 2022.
\newblock \href {https://doi.org/10.1109/ACCESS.2022.3196347} {Code generation using machine learning: {A} systematic review}.
\newblock \emph{{IEEE} Access}, 10:82434--82455.

\bibitem[{Finlayson et~al.(2024)Finlayson, Hewitt, Koller, Swayamdipta, and Sabharwal}]{DBLP:conf/iclr/FinlaysonHKSS24}
Matthew Finlayson, John Hewitt, Alexander Koller, Swabha Swayamdipta, and Ashish Sabharwal. 2024.
\newblock \href {https://openreview.net/forum?id=dONpC9GL1o} {Closing the curious case of neural text degeneration}.
\newblock In \emph{The Twelfth International Conference on Learning Representations, {ICLR} 2024, Vienna, Austria, May 7-11, 2024}. OpenReview.net.

\bibitem[{Ganaie et~al.(2022)Ganaie, Hu, Malik, Tanveer, and Suganthan}]{DBLP:journals/eaai/GanaieHMTS22}
M.~A. Ganaie, Minghui Hu, Ashwani~Kumar Malik, Muhammad Tanveer, and Ponnuthurai~N. Suganthan. 2022.
\newblock \href {https://doi.org/10.1016/J.ENGAPPAI.2022.105151} {Ensemble deep learning: {A} review}.
\newblock \emph{Eng. Appl. Artif. Intell.}, 115:105151.

\bibitem[{Gu et~al.(2018)Gu, Bradbury, Xiong, Li, and Socher}]{DBLP:conf/iclr/Gu0XLS18}
Jiatao Gu, James Bradbury, Caiming Xiong, Victor O.~K. Li, and Richard Socher. 2018.
\newblock \href {https://openreview.net/forum?id=B1l8BtlCb} {Non-autoregressive neural machine translation}.
\newblock In \emph{6th International Conference on Learning Representations, {ICLR} 2018, Vancouver, BC, Canada, April 30 - May 3, 2018, Conference Track Proceedings}. OpenReview.net.

\bibitem[{Heineman et~al.(2024)Heineman, Dou, and Xu}]{DBLP:conf/emnlp/HeinemanD024}
David Heineman, Yao Dou, and Wei Xu. 2024.
\newblock \href {https://aclanthology.org/2024.emnlp-main.1255} {Improving minimum bayes risk decoding with multi-prompt}.
\newblock In \emph{Proceedings of the 2024 Conference on Empirical Methods in Natural Language Processing, {EMNLP} 2024, Miami, FL, USA, November 12-16, 2024}, pages 22525--22545. Association for Computational Linguistics.

\bibitem[{Hokamp et~al.(2020)Hokamp, Ghalandari, Pham, and Glover}]{DBLP:journals/corr/abs-2006-08748}
Chris Hokamp, Demian~Gholipour Ghalandari, Nghia~The Pham, and John Glover. 2020.
\newblock \href {https://arxiv.org/abs/2006.08748} {Dyne: Dynamic ensemble decoding for multi-document summarization}.
\newblock \emph{CoRR}, abs/2006.08748.

\bibitem[{Jiang et~al.(2024)Jiang, Wang, Shen, Kim, and Kim}]{DBLP:journals/corr/abs-2406-00515}
Juyong Jiang, Fan Wang, Jiasi Shen, Sungju Kim, and Sunghun Kim. 2024.
\newblock \href {https://doi.org/10.48550/ARXIV.2406.00515} {A survey on large language models for code generation}.
\newblock \emph{CoRR}, abs/2406.00515.

\bibitem[{Jiang et~al.(2020)Jiang, Xu, Araki, and Neubig}]{DBLP:journals/tacl/JiangXAN20}
Zhengbao Jiang, Frank~F. Xu, Jun Araki, and Graham Neubig. 2020.
\newblock \href {https://doi.org/10.1162/TACL\_A\_00324} {How can we know what language models know}.
\newblock \emph{Trans. Assoc. Comput. Linguistics}, 8:423--438.

\bibitem[{Lakshminarayanan et~al.(2017)Lakshminarayanan, Pritzel, and Blundell}]{DBLP:conf/nips/Lakshminarayanan17}
Balaji Lakshminarayanan, Alexander Pritzel, and Charles Blundell. 2017.
\newblock \href {https://proceedings.neurips.cc/paper/2017/hash/9ef2ed4b7fd2c810847ffa5fa85bce38-Abstract.html} {Simple and scalable predictive uncertainty estimation using deep ensembles}.
\newblock In \emph{Advances in Neural Information Processing Systems 30: Annual Conference on Neural Information Processing Systems 2017, December 4-9, 2017, Long Beach, CA, {USA}}, pages 6402--6413.

\bibitem[{Liu et~al.(2023)Liu, Yuan, Fu, Jiang, Hayashi, and Neubig}]{DBLP:journals/csur/LiuYFJHN23}
Pengfei Liu, Weizhe Yuan, Jinlan Fu, Zhengbao Jiang, Hiroaki Hayashi, and Graham Neubig. 2023.
\newblock \href {https://doi.org/10.1145/3560815} {Pre-train, prompt, and predict: {A} systematic survey of prompting methods in natural language processing}.
\newblock \emph{{ACM} Comput. Surv.}, 55(9):195:1--195:35.

\bibitem[{Maddela et~al.(2023)Maddela, Dou, Heineman, and Xu}]{DBLP:conf/acl/MaddelaDHX23}
Mounica Maddela, Yao Dou, David Heineman, and Wei Xu. 2023.
\newblock \href {https://doi.org/10.18653/V1/2023.ACL-LONG.905} {{LENS:} {A} learnable evaluation metric for text simplification}.
\newblock In \emph{Proceedings of the 61st Annual Meeting of the Association for Computational Linguistics (Volume 1: Long Papers), {ACL} 2023, Toronto, Canada, July 9-14, 2023}, pages 16383--16408. Association for Computational Linguistics.

\bibitem[{Nakamachi et~al.(2020)Nakamachi, Kajiwara, and Arase}]{DBLP:conf/ijcnlp/NakamachiKA20}
Akifumi Nakamachi, Tomoyuki Kajiwara, and Yuki Arase. 2020.
\newblock \href {https://aclanthology.org/2020.aacl-srw.22/} {Text simplification with reinforcement learning using supervised rewards on grammaticality, meaning preservation, and simplicity}.
\newblock In \emph{Proceedings of the 1st Conference of the Asia-Pacific Chapter of the Association for Computational Linguistics and the 10th International Joint Conference on Natural Language Processing: Student Research Workshop, {AACL/IJCNLP} 2021, Suzhou, China, December 4-7, 2020}, pages 153--159. Association for Computational Linguistics.

\bibitem[{OpenAI(2023)}]{DBLP:journals/corr/abs-2303-08774}
OpenAI. 2023.
\newblock \href {https://doi.org/10.48550/ARXIV.2303.08774} {{GPT-4} technical report}.
\newblock \emph{CoRR}, abs/2303.08774.

\bibitem[{Papineni et~al.(2002)Papineni, Roukos, Ward, and Zhu}]{DBLP:conf/acl/PapineniRWZ02}
Kishore Papineni, Salim Roukos, Todd Ward, and Wei{-}Jing Zhu. 2002.
\newblock \href {https://doi.org/10.3115/1073083.1073135} {Bleu: a method for automatic evaluation of machine translation}.
\newblock In \emph{Proceedings of the 40th Annual Meeting of the Association for Computational Linguistics, July 6-12, 2002, Philadelphia, PA, {USA}}, pages 311--318. {ACL}.

\bibitem[{Pitis et~al.(2023)Pitis, Zhang, Wang, and Ba}]{DBLP:journals/corr/abs-2304-05970}
Silviu Pitis, Michael~R. Zhang, Andrew Wang, and Jimmy Ba. 2023.
\newblock \href {https://doi.org/10.48550/ARXIV.2304.05970} {Boosted prompt ensembles for large language models}.
\newblock \emph{CoRR}, abs/2304.05970.

\bibitem[{Rei et~al.(2020)Rei, Stewart, Farinha, and Lavie}]{DBLP:conf/emnlp/ReiSFL20}
Ricardo Rei, Craig Stewart, Ana~C. Farinha, and Alon Lavie. 2020.
\newblock \href {https://doi.org/10.18653/V1/2020.EMNLP-MAIN.213} {{COMET:} {A} neural framework for {MT} evaluation}.
\newblock In \emph{Proceedings of the 2020 Conference on Empirical Methods in Natural Language Processing, {EMNLP} 2020, Online, November 16-20, 2020}, pages 2685--2702. Association for Computational Linguistics.

\bibitem[{Sennrich et~al.(2016)Sennrich, Haddow, and Birch}]{sennrich2016improving}
Rico Sennrich, Barry Haddow, and Alexandra Birch. 2016.
\newblock \href {https://arxiv.org/abs/1511.06709} {Improving neural machine translation models with monolingual data}.
\newblock \emph{Preprint}, arXiv:1511.06709.

\bibitem[{Touvron et~al.(2023{\natexlab{a}})Touvron, Lavril, Izacard, Martinet, Lachaux, Lacroix, Rozi{\`{e}}re, Goyal, Hambro, Azhar, Rodriguez, Joulin, Grave, and Lample}]{DBLP:journals/corr/abs-2302-13971}
Hugo Touvron, Thibaut Lavril, Gautier Izacard, Xavier Martinet, Marie{-}Anne Lachaux, Timoth{\'{e}}e Lacroix, Baptiste Rozi{\`{e}}re, Naman Goyal, Eric Hambro, Faisal Azhar, Aur{\'{e}}lien Rodriguez, Armand Joulin, Edouard Grave, and Guillaume Lample. 2023{\natexlab{a}}.
\newblock \href {https://doi.org/10.48550/ARXIV.2302.13971} {Llama: Open and efficient foundation language models}.
\newblock \emph{CoRR}, abs/2302.13971.

\bibitem[{Touvron et~al.(2023{\natexlab{b}})Touvron, Martin, Stone, Albert, Almahairi, Babaei, Bashlykov, Batra, Bhargava, Bhosale, Bikel, Blecher, Canton{-}Ferrer, Chen, Cucurull, Esiobu, Fernandes, Fu, Fu, Fuller, Gao, Goswami, Goyal, Hartshorn, Hosseini, Hou, Inan, Kardas, Kerkez, Khabsa, Kloumann, Korenev, Koura, Lachaux, Lavril, Lee, Liskovich, Lu, Mao, Martinet, Mihaylov, Mishra, Molybog, Nie, Poulton, Reizenstein, Rungta, Saladi, Schelten, Silva, Smith, Subramanian, Tan, Tang, Taylor, Williams, Kuan, Xu, Yan, Zarov, Zhang, Fan, Kambadur, Narang, Rodriguez, Stojnic, Edunov, and Scialom}]{DBLP:journals/corr/abs-2307-09288}
Hugo Touvron, Louis Martin, Kevin Stone, Peter Albert, Amjad Almahairi, Yasmine Babaei, Nikolay Bashlykov, Soumya Batra, Prajjwal Bhargava, Shruti Bhosale, Dan Bikel, Lukas Blecher, Cristian Canton{-}Ferrer, Moya Chen, Guillem Cucurull, David Esiobu, Jude Fernandes, Jeremy Fu, Wenyin Fu, Brian Fuller, Cynthia Gao, Vedanuj Goswami, Naman Goyal, Anthony Hartshorn, Saghar Hosseini, Rui Hou, Hakan Inan, Marcin Kardas, Viktor Kerkez, Madian Khabsa, Isabel Kloumann, Artem Korenev, Punit~Singh Koura, Marie{-}Anne Lachaux, Thibaut Lavril, Jenya Lee, Diana Liskovich, Yinghai Lu, Yuning Mao, Xavier Martinet, Todor Mihaylov, Pushkar Mishra, Igor Molybog, Yixin Nie, Andrew Poulton, Jeremy Reizenstein, Rashi Rungta, Kalyan Saladi, Alan Schelten, Ruan Silva, Eric~Michael Smith, Ranjan Subramanian, Xiaoqing~Ellen Tan, Binh Tang, Ross Taylor, Adina Williams, Jian~Xiang Kuan, Puxin Xu, Zheng Yan, Iliyan Zarov, Yuchen Zhang, Angela Fan, Melanie Kambadur, Sharan Narang, Aur{\'{e}}lien Rodriguez, Robert Stojnic, Sergey Edunov,
  and Thomas Scialom. 2023{\natexlab{b}}.
\newblock \href {https://doi.org/10.48550/ARXIV.2307.09288} {Llama 2: Open foundation and fine-tuned chat models}.
\newblock \emph{CoRR}, abs/2307.09288.

\bibitem[{Vaswani et~al.(2017)Vaswani, Shazeer, Parmar, Uszkoreit, Jones, Gomez, Kaiser, and Polosukhin}]{DBLP:conf/nips/VaswaniSPUJGKP17}
Ashish Vaswani, Noam Shazeer, Niki Parmar, Jakob Uszkoreit, Llion Jones, Aidan~N. Gomez, Lukasz Kaiser, and Illia Polosukhin. 2017.
\newblock \href {https://proceedings.neurips.cc/paper/2017/hash/3f5ee243547dee91fbd053c1c4a845aa-Abstract.html} {Attention is all you need}.
\newblock In \emph{Advances in Neural Information Processing Systems 30: Annual Conference on Neural Information Processing Systems 2017, December 4-9, 2017, Long Beach, CA, {USA}}, pages 5998--6008.

\bibitem[{Wang et~al.(2023)Wang, Wei, Schuurmans, Le, Chi, Narang, Chowdhery, and Zhou}]{DBLP:conf/iclr/0002WSLCNCZ23}
Xuezhi Wang, Jason Wei, Dale Schuurmans, Quoc~V. Le, Ed~H. Chi, Sharan Narang, Aakanksha Chowdhery, and Denny Zhou. 2023.
\newblock \href {https://openreview.net/forum?id=1PL1NIMMrw} {Self-consistency improves chain of thought reasoning in language models}.
\newblock In \emph{The Eleventh International Conference on Learning Representations, {ICLR} 2023, Kigali, Rwanda, May 1-5, 2023}. OpenReview.net.

\bibitem[{Wei et~al.(2023)Wei, Wu, Shang, Li, Wang, Guo, Chen, Yu, and Yang}]{wei2023text}
Daimeng Wei, Zhanglin Wu, Hengchao Shang, Zongyao Li, Minghan Wang, Jiaxin Guo, Xiaoyu Chen, Zhengzhe Yu, and Hao Yang. 2023.
\newblock \href {https://arxiv.org/abs/2306.01318} {Text style transfer back-translation}.
\newblock \emph{Preprint}, arXiv:2306.01318.

\bibitem[{Wei et~al.(2022)Wei, Wang, Schuurmans, Bosma, Ichter, Xia, Chi, Le, and Zhou}]{DBLP:conf/nips/Wei0SBIXCLZ22}
Jason Wei, Xuezhi Wang, Dale Schuurmans, Maarten Bosma, Brian Ichter, Fei Xia, Ed~H. Chi, Quoc~V. Le, and Denny Zhou. 2022.
\newblock \href {http://papers.nips.cc/paper\_files/paper/2022/hash/9d5609613524ecf4f15af0f7b31abca4-Abstract-Conference.html} {Chain-of-thought prompting elicits reasoning in large language models}.
\newblock In \emph{Advances in Neural Information Processing Systems 35: Annual Conference on Neural Information Processing Systems 2022, NeurIPS 2022, New Orleans, LA, USA, November 28 - December 9, 2022}.

\bibitem[{Xie et~al.(2020)Xie, Dai, Chen, Dai, Zhao, Zha, Wei, and Pfister}]{DBLP:conf/nips/XieDCDZZ0P20}
Yujia Xie, Hanjun Dai, Minshuo Chen, Bo~Dai, Tuo Zhao, Hongyuan Zha, Wei Wei, and Tomas Pfister. 2020.
\newblock \href {https://proceedings.neurips.cc/paper/2020/hash/ec24a54d62ce57ba93a531b460fa8d18-Abstract.html} {Differentiable top-k with optimal transport}.
\newblock In \emph{Advances in Neural Information Processing Systems 33: Annual Conference on Neural Information Processing Systems 2020, NeurIPS 2020, December 6-12, 2020, virtual}.

\bibitem[{Yang et~al.(2024)Yang, Yang, Hui, Zheng, Yu, Zhou, Li, Li, Liu, Huang, Dong, Wei, Lin, Tang, Wang, Yang, Tu, Zhang, Ma, Yang, Xu, Zhou, Bai, He, Lin, Dang, Lu, Chen, Yang, Li, Xue, Ni, Zhang, Wang, Peng, Men, Gao, Lin, Wang, Bai, Tan, Zhu, Li, Liu, Ge, Deng, Zhou, Ren, Zhang, Wei, Ren, Liu, Fan, Yao, Zhang, Wan, Chu, Liu, Cui, Zhang, Guo, and Fan}]{DBLP:journals/corr/abs-2407-10671}
An~Yang, Baosong Yang, Binyuan Hui, Bo~Zheng, Bowen Yu, Chang Zhou, Chengpeng Li, Chengyuan Li, Dayiheng Liu, Fei Huang, Guanting Dong, Haoran Wei, Huan Lin, Jialong Tang, Jialin Wang, Jian Yang, Jianhong Tu, Jianwei Zhang, Jianxin Ma, Jianxin Yang, Jin Xu, Jingren Zhou, Jinze Bai, Jinzheng He, Junyang Lin, Kai Dang, Keming Lu, Keqin Chen, Kexin Yang, Mei Li, Mingfeng Xue, Na~Ni, Pei Zhang, Peng Wang, Ru~Peng, Rui Men, Ruize Gao, Runji Lin, Shijie Wang, Shuai Bai, Sinan Tan, Tianhang Zhu, Tianhao Li, Tianyu Liu, Wenbin Ge, Xiaodong Deng, Xiaohuan Zhou, Xingzhang Ren, Xinyu Zhang, Xipin Wei, Xuancheng Ren, Xuejing Liu, Yang Fan, Yang Yao, Yichang Zhang, Yu~Wan, Yunfei Chu, Yuqiong Liu, Zeyu Cui, Zhenru Zhang, Zhifang Guo, and Zhihao Fan. 2024.
\newblock \href {https://doi.org/10.48550/ARXIV.2407.10671} {Qwen2 technical report}.
\newblock \emph{CoRR}, abs/2407.10671.

\bibitem[{Zhang et~al.(2020)Zhang, Kishore, Wu, Weinberger, and Artzi}]{DBLP:conf/iclr/ZhangKWWA20}
Tianyi Zhang, Varsha Kishore, Felix Wu, Kilian~Q. Weinberger, and Yoav Artzi. 2020.
\newblock \href {https://openreview.net/forum?id=SkeHuCVFDr} {Bertscore: Evaluating text generation with {BERT}}.
\newblock In \emph{8th International Conference on Learning Representations, {ICLR} 2020, Addis Ababa, Ethiopia, April 26-30, 2020}. OpenReview.net.

\bibitem[{Zhao et~al.(2023)Zhao, Wang, and Yang}]{DBLP:conf/ijcai/ZhaoWY23}
Jiangjiang Zhao, Zhuoran Wang, and Fangchun Yang. 2023.
\newblock \href {https://doi.org/10.24963/IJCAI.2023/588} {Genetic prompt search via exploiting language model probabilities}.
\newblock In \emph{Proceedings of the Thirty-Second International Joint Conference on Artificial Intelligence, {IJCAI} 2023, 19th-25th August 2023, Macao, SAR, China}, pages 5296--5305. ijcai.org.

\bibitem[{Zhou et~al.(2002)Zhou, Wu, and Tang}]{DBLP:journals/ai/ZhouWT02}
Zhi{-}Hua Zhou, Jianxin Wu, and Wei Tang. 2002.
\newblock \href {https://doi.org/10.1016/S0004-3702(02)00190-X} {Ensembling neural networks: Many could be better than all}.
\newblock \emph{Artif. Intell.}, 137(1-2):239--263.

\end{thebibliography}

\clearpage
\onecolumn
\appendix
\section{Prompts Design}
\label{sec:appendix_prompt}

\subsection{Prompts for Machine Translation Task}
\label{sec:appendix_prompt_mt}

The designed prompts for Machine Translation Task used in Section \ref{sec:mt_task} are as following:

\begin{table*}[htbp]
\centering
\begin{tabularx}{\textwidth}{X}
\hline
\multicolumn{1}{c}{$p_1$} \\
\hdashline
\textit{\textbf{System Prompt:}} \\
You are a great translation assistant! \\
\\
\textit{\textbf{User Prompt:}} \\
Translate the following paragraph from \{\textit{source language}\} to \{\textit{target language}\}, ensuring that no part of the sentence is omitted. \\
\{\textit{source language}\}: \{\textit{source text}\} \\
\hline
\multicolumn{1}{c}{$p_2$} \\
\hdashline
\textit{\textbf{System Prompt:}} \\
You are a helpful assistant! \\
\\
\textit{\textbf{User Prompt:}} \\
You're a very professional translator. Please help me translate the following paragraph from \{\textit{source language}\} to \{\textit{target language}\}. \\
\{\textit{source language}\}: \{\textit{source text}\} \\
\hline
\end{tabularx}
\caption{Designed Prompts for Machine Translation Task.}
\label{table:appendix_prompt_mt}
\end{table*}

\clearpage
\subsection{Prompts for Code Generation Task}
\label{sec:appendix_prompt_code}

For Code Generation Task, since the dataset provides a prompt for each problem, we define this original prompt as $p_1$. For prompt $p_2$, we construct a simple prefix based on $p_1$. Examples of this prompts used in Section \ref{sec:code_task} are as following:

\begin{table*}[htbp]
\centering
\begin{tabularx}{\textwidth}{X}
\hline
\multicolumn{1}{c}{$p_1$} \\
\hdashline
def specialFilter(nums): \\
\ \ \ \ \ """Write a function that takes an array of numbers as input and returns the number of elements in the array that are greater than 10 and both first and last digits of a number are odd (1, 3, 5, 7, 9). \\
\ \ \ \ \ For example: \\
\ \ \ \ \ specialFilter([15, -73, 14, -15]) => 1 \\
\ \ \ \ \ specialFilter([33, -2, -3, 45, 21, 109]) => 2 \\
\ \ \ \ \ """ \\
\hline
\multicolumn{1}{c}{$p_2$} \\
\hdashline
\textbf{""" This is a good code.} \\
def specialFilter(nums): \\
\ \ \ \ \ """Write a function that takes an array of numbers as input and returns the number of elements in the array that are greater than 10 and both first and last digits of a number are odd (1, 3, 5, 7, 9). \\
\ \ \ \ \ For example: \\
\ \ \ \ \ specialFilter([15, -73, 14, -15]) => 1 \\
\ \ \ \ \ specialFilter([33, -2, -3, 45, 21, 109]) => 2 \\
\ \ \ \ \ """ \\
\hline
\end{tabularx}
\caption{Prompts of ID 146 for Code Generation Task}
\label{table:appendix_prompt_code_example1}
\end{table*}

\begin{table*}[!h]
\centering
\begin{tabularx}{\textwidth}{X}
\hline
\multicolumn{1}{c}{$p_1$} \\
\hdashline
from typing import List, Tuple \\
\\
def rolling\_max(numbers: List[int]) -> List[int]: \\
\ \ \ \ \ """ From a given list of integers, generate a list of rolling maximum element found until given moment in the sequence. \\
\ \ \ \ \ \>\>\> rolling\_max([1, 2, 3, 2, 3, 4, 2]) \\
\ \ \ \ \ [1, 2, 3, 3, 3, 4, 4] \\
\ \ \ \ \ """ \\
\hline
\multicolumn{1}{c}{$p_2$} \\
\hdashline
\textbf{""" This is a good code.} \\
from typing import List, Tuple \\
\\
def rolling\_max(numbers: List[int]) -> List[int]: \\
\ \ \ \ \ """ From a given list of integers, generate a list of rolling maximum element found until given moment in the sequence. \\
\ \ \ \ \ \>\>\> rolling\_max([1, 2, 3, 2, 3, 4, 2]) \\
\ \ \ \ \ [1, 2, 3, 3, 3, 4, 4] \\
\ \ \ \ \ """ \\
\hline
\end{tabularx}
\caption{Prompts of ID 9 for Code Generation Task}
\label{table:appendix_prompt_code_example2}
\end{table*}

\clearpage
\subsection{Prompts for Text Simplification Task}
\label{sec:appendix_prompt_ts}

The designed prompts for Text Simplification used in Section \ref{sec:ts_task} are as following:

\begin{table*}[!h]
\centering
\begin{tabularx}{\textwidth}{X}
\hline
\multicolumn{1}{c}{$p_1$} \\
\hdashline
\textit{\textbf{User Prompt:}} \\
Please simplify the following sentence so that it is easy to understand by people with disabilities or those who are unfamiliar with English. Try to use shorter words, fewer clauses, and a simpler structure. \\
Original: \{\textit{input text}\} \\
\hline
\multicolumn{1}{c}{$p_2$} \\
\hdashline
\textit{\textbf{User Prompt:}} \\
Create a simpler version of the sentence below so that it can be better understood by non-English speakers or individuals with disabilities. \\
Original: \{\textit{input text}\} \\
\hline
\end{tabularx}
\caption{Designed Prompts for Text Simplification Task}
\label{table:appendix_prompt_ts}
\end{table*}

\clearpage
\section{Detailed Main Experimental Results}
\label{sec:appendix_rsts}
\subsection{Detailed Results for Machine Translation Task}
\label{sec:appendix_rsts_mt}

% The detailed results upon different seed conditions for Machine Translation Task used in Section \ref{sec:mt_task} are as following:
The specific outcomes for various seed settings in the Machine Translation Task, as discussed in Section \ref{sec:mt_task}, are presented below in Table \ref{table:appendix_rsts_mt}:

\begin{table*}[!h]
\centering
\begin{tabularx}{\textwidth}{Xcccccccc}
\hline
% \toprule[1.1pt]
$d$-BLEU & \textbf{en$\rightarrow$de} & \textbf{en$\rightarrow$fr} & \textbf{en$\rightarrow$zh} & \textbf{en$\rightarrow$ja} & \textbf{de$\rightarrow$en} & \textbf{fr$\rightarrow$en} & \textbf{zh$\rightarrow$en} & \textbf{ja$\rightarrow$en} \\
\hline
 & \multicolumn{8}{c}{$p_1$}\\
\hdashline
0 & 27.18 & 38.57 & 22.53 & 13.36 & 34.74 & 42.27 & 21.95 & 9.21 \\
1 & 27.82 & 37.6 & 16.43 & 7.23 & 34.87 & 42.22 & 25.84 & 7.26 \\
2 & 27.4 & 35.94 & 16.58 & 5.52 & 31.78 & 39.12 & 24.3 & 5.58 \\
3 & 28.2 & 37.6 & 15.53 & 12.08 & 30.99 & 41.81 & 25.06 & 5.58 \\
4 & 26.54 & 37.88 & 16.93 & 11.7 & 32.02 & 41.96 & 24.31 & 4.69 \\
5 & 27.2 & 36.83 & 15.61 & 7.87 & 32.22 & 41.88 & 21.9 & 9.31 \\
6 & 27.1 & 38.43 & 15.79 & 6.85 & 31.19 & 41.17 & 25.17 & 6.95 \\
7 & 28 & 37.79 & 22.44 & 8.47 & 34.92 & 42.36 & 24.82 & 5.21 \\
8 & 26.69 & 37.61 & 15.56 & 14.02 & 32.45 & 39.5 & 26.24 & 6.88 \\
9 & 27.86 & 37.33 & 16.3 & 6.39 & 33.7 & 41.49 & 24.35 & 6.51 \\
\hdashline
AVG & 27.40 & 37.59 & 17.37 & 9.35 & 32.89 & 41.38 & 24.39 & 6.72 \\
\hline
 & \multicolumn{8}{c}{$p_2$}\\
\hdashline
0 & 26.99 & 35.85 & 15.14 & 14.03 & 35.21 & 41.88 & 24.94 & 6.76 \\
1 & 26.04 & 37.68 & 16.74 & 12.29 & 33.84 & 42.63 & 24.63 & 6.76 \\
2 & 26.57 & 38.5 & 16.09 & 6.19 & 31.73 & 42.61 & 26.39 & 5.1 \\
3 & 26.15 & 37.12 & 16.47 & 6.46 & 34.89 & 42.42 & 21.68 & 5.2 \\
4 & 27.08 & 37.61 & 21.57 & 14.14 & 31.29 & 39.37 & 22.02 & 6.87 \\
5 & 26.36 & 37.01 & 14.85 & 7.73 & 30.28 & 41.8 & 24.49 & 9.45 \\
6 & 26.96 & 36.59 & 16.65 & 12.3 & 31.61 & 41.87 & 24.75 & 5.13 \\
7 & 27.9 & 35.82 & 21.43 & 6.37 & 31.74 & 42.06 & 25.17 & 5.24 \\
8 & 27.61 & 37.24 & 17.38 & 8.82 & 34.81 & 42.3 & 25.28 & 6.76 \\
9 & 28.79 & 37.97 & 16 & 7.81 & 31.97 & 39.01 & 24.5 & 10.25 \\
\hdashline
AVG & 27.05 & 37.14 & 17.23 & 9.61 & 32.74 & 41.60 & 24.39 & 6.75 \\
\hline
 & \multicolumn{8}{c}{Ours}\\
\hdashline
0 & 28.58 & 37.79 & 18.86 & 9.01 & 32.01 & 43.21 & 26.06 & 7.93 \\
1 & 27.66 & 38.84 & 16.23 & 14.94 & 34.74 & 42.8 & 26.13 & 7.74 \\
2 & 28.59 & 39.04 & 23.06 & 11.06 & 34.79 & 42.86 & 25.18 & 10.79 \\
3 & 28.29 & 39.15 & 18.75 & 12.71 & 35.15 & 39.87 & 26.09 & 6.82 \\
4 & 27.55 & 37.94 & 17.82 & 14.66 & 33.4 & 42.83 & 25.61 & 7.35 \\
5 & 28.82 & 36.84 & 17.02 & 8.87 & 32.61 & 42.71 & 24.66 & 7.95 \\
6 & 26.66 & 37.96 & 22.83 & 12.74 & 31.86 & 42.74 & 21.98 & 7.98 \\
7 & 26.59 & 39.16 & 19.08 & 9.56 & 32.18 & 42.75 & 23.81 & 10.9 \\
8 & 27.97 & 37.19 & 18.9 & 9.89 & 32.61 & 39.69 & 22.92 & 8.08 \\
9 & 28.07 & 37.72 & 16.29 & 10.91 & 31.73 & 43.06 & 26.04 & 8.11 \\
\hdashline
AVG & 27.88 & 38.16 & 18.38 & 11.44 & 33.11 & 42.25 & 24.85 & 8.37 \\
\hline
% \bottomrule[1.2pt]
\end{tabularx}
\caption{Detailed results for Machine Translation Task.}
\label{table:appendix_rsts_mt}
\end{table*}

\clearpage
\twocolumn

\subsection{Detailed Results for Code Generation Task}
\label{sec:appendix_rsts_code}

The specific outcomes for various seed settings in the Code Generation Task, as discussed in Section \ref{sec:code_task}, are presented below in Table \ref{table:appendix_rsts_code}:

\begin{table}[h]
\centering
\begin{tabularx}{0.48\textwidth}{Xccc}
\hline
\textbf{pass@$k$} & \textbf{$k$=1} & \textbf{$k$=5} & \textbf{$k$=10} \\
\hline
& \multicolumn{3}{c}{$p_1$} \\
\hdashline
0 & 30.41\% & 56.78\% & 67.67\% \\
1 & 31.94\% & 58.73\% & 66.68\% \\
2 & 32.32\% & 58.03\% & 67.92\% \\
3 & 39.24\% & 59.19\% & 66.39\% \\
4 & 30.44\% & 59.17\% & 67.42\% \\
5 & 39.2\% & 58.97\% & 66.32\% \\
6 & 28.02\% & 57.13\% & 67.72\% \\
7 & 30.71\% & 57.81\% & 67.53\% \\
8 & 28.42\% & 57.63\% & 67.76\% \\
9 & 30.37\% & 57.88\% & 66.33\% \\
\hdashline
AVG & 32.11\% & 58.13\% & 67.17\% \\
\hline
& \multicolumn{3}{c}{$p_2$} \\
\hdashline
0 & 30.11\% & 58.64\% & 67.34\% \\
1 & 32.12\% & 58.73\% & 67.13\% \\
2 & 32.31\% & 57.62\% & 67.88\% \\
3 & 33.39\% & 58.97\% & 67.41\% \\
4 & 28.05\% & 56.39\% & 67\% \\
5 & 33.1\% & 57.9\% & 66.92\% \\
6 & 30.1\% & 58.58\% & 67.71\% \\
7 & 27.95\% & 58.47\% & 66.69\% \\
8 & 30.35\% & 58.67\% & 67.16\% \\
9 & 29.96\% & 56.3\% & 67.35\% \\
\hdashline
AVG & 30.74\% & 58.03\% & 67.26\% \\
\hline
& \multicolumn{3}{c}{Ours} \\
\hdashline
0 & 32.66\% & 58.9\% & 67.2\% \\
1 & 37.1\% & 58.86\% & 69.12\% \\
2 & 37.06\% & 57.89\% & 67.68\% \\
3 & 32.97\% & 60.05\% & 67.76\% \\
4 & 31.8\% & 58.9\% & 67.27\% \\
5 & 33.03\% & 59.21\% & 68.5\% \\
6 & 32.16\% & 59.19\% & 68.05\% \\
7 & 33.26\% & 57.93\% & 69.27\% \\
8 & 30.59\% & 59.37\% & 67.62\% \\
9 & 30.57\% & 60.41\% & 68.62\% \\
\hdashline
AVG & 33.12\% & 59.07\% & 68.11\% \\
\hline
\end{tabularx}
\caption{Detailed Results for Code Generation Task.}
\label{table:appendix_rsts_code}
\end{table}

\newpage
% \columnbreak
\subsection{Detailed Results for Text Simplification Task}
\label{sec:appendix_rsts_ts}

The specific outcomes for various seed settings in the Text Simplification Task, as discussed in Section \ref{sec:ts_task}, are presented below in Table \ref{table:appendix_rsts_ts}:

\begin{table}[htbp]
\centering
\begin{tabularx}{0.48\textwidth}{Xcccc}
\hline
\textbf{LENS} & \textbf{$w/$ ref} & & \textbf{$w/o$ ref} & \\
\hline
& \multicolumn{4}{c}{$p_1$} \\
\hdashline
0 & 75.32 &  & 80.76 & \\
1 & 75.93 &  & 82.04 & \\
2 & 75.65 &  & 82.58 & \\
3 & 74.47 &  & 81.02 & \\
4 & 75.71 &  & 82.2 & \\
5 & 74.43 &  & 81.4 & \\
6 & 75.37 &  & 82.36 & \\
7 & 75.36 &  & 80.26 & \\
8 & 74.59 &  & 82.5 & \\
9 & 73.98 &  & 80.27 & \\
\hdashline
AVG & 75.08 &  & 81.54 & \\
\hline
& \multicolumn{4}{c}{$p_2$} \\
\hdashline
0 & 74.54 &  & 81.79 & \\
1 & 75.81 &  & 80.62 & \\
2 & 72.83 &  & 82.44 & \\
3 & 74.48 &  & 81.61 & \\
4 & 73.31 &  & 82.36 & \\
5 & 75.68 &  & 80.91 & \\
6 & 74.92 &  & 81.65 & \\
7 & 75.35 &  & 81.23 & \\
8 & 74.63 &  & 81.7 & \\
9 & 76.05 &  & 81.99 & \\
\hdashline
AVG & 74.76 &  & 81.63 & \\
\hline
& \multicolumn{4}{c}{Ours} \\
\hdashline
0 & 76.71 &  & 82.21 & \\
1 & 76.03 &  & 81.88 & \\
2 & 77.26 &  & 81.68 & \\
3 & 77.22 &  & 81.98 & \\
4 & 77.38 &  & 82.36 & \\
5 & 78.69 &  & 82.61 & \\
6 & 78.02 &  & 82.12 & \\
7 & 76.17 &  & 81.63 & \\
8 & 77.96 &  & 82.64 & \\
9 & 76.32 &  & 81.69 & \\
\hdashline
AVG & 77.18 &  & 82.08 & \\
\hline
\end{tabularx}
\caption{Results for Text Simplification Task.}
\label{table:appendix_rsts_ts}
\end{table}

\clearpage
\section{Appendix for Study}
\label{sec:appendix_study}

\subsection{Detailed Results for Study under Various Decoding Strategies}
\label{sec:appendix_study_decoding}

The specific outcomes for various seed settings in the Machine Translation Task under Top-$k$ decoding strategy, as discussed in Section \ref{sec:study_decoding}, are presented below in Table \ref{table:appendix_study_decoding_mt}:

\begin{table}[!h]
\centering
\begin{tabularx}{0.48\textwidth}{Xcc}
\hline
% \toprule[1.1pt]
$d$-BLEU & \textbf{en$\rightarrow$de} & \textbf{en$\rightarrow$zh} \\
\hline
\multicolumn{3}{c}{$p_1$} \\
\hdashline
0 & 25.88 & 21.43 \\
1 & 27.92 & 15.99 \\
2 & 26.15 & 15.62 \\
3 & 26.99 & 20.85 \\
4 & 26.03 & 16.34 \\
5 & 27.73 & 14.90 \\
6 & 28.04 & 16.12 \\
7 & 26.80 & 15.73 \\
8 & 26.71 & 22.51 \\
9 & 26.90 & 16.15 \\
\hdashline
AVG & 26.92 & 17.56 \\
\hline
\multicolumn{3}{c}{$p_2$} \\
\hdashline
0 & 28.79 & 17.70 \\
1 & 26.60 & 16.99 \\
2 & 28.26 & 16.55 \\
3 & 27.28 & 15.37 \\
4 & 26.77 & 16.67 \\
5 & 27.82 & 15.20 \\
6 & 27.16 & 14.88 \\
7 & 25.64 & 16.15 \\
8 & 27.56 & 15.51 \\
9 & 26.36 & 21.94 \\
\hdashline
AVG & 27.22 & 16.70 \\
\hline
\multicolumn{3}{c}{Ours} \\
\hdashline
0 & 28.60 & 18.79 \\
1 & 28.24 & 18.57 \\
2 & 25.54 & 17.61 \\
3 & 27.49 & 19.43 \\
4 & 25.92 & 16.40 \\
5 & 26.75 & 19.76 \\
6 & 28.66 & 17.57 \\
7 & 27.93 & 17.10 \\
8 & 28.58 & 22.24 \\
9 & 27.46 & 21.20 \\
\hdashline
AVG & 27.52 & 18.87 \\
% \bottomrule[1.2pt]
\end{tabularx}
\caption{Detailed results for Machine Translation Task under Top-$k$ decoding strategy.}
\label{table:appendix_study_decoding_mt}
\end{table}

\newpage
The specific outcomes for various seed settings in the Code Generation Task under Top-$k$ decoding strategy, as discussed in Section \ref{sec:study_decoding}, are presented below in Table \ref{table:appendix_study_decoding_code}:

\begin{table}[!h]
\centering
\begin{tabularx}{0.48\textwidth}{Xc}
\hline
% \toprule[1.1pt]
pass-$k$ & $k$=10 \\
\hline
\multicolumn{2}{c}{$p_1$} \\
\hdashline
0 & 66.99\% \\
1 & 66.31\% \\
2 & 67.57\% \\
3 & 66.25\% \\
4 & 67.25\% \\
5 & 65.83\% \\
6 & 67.66\% \\
7 & 67.84\% \\
8 & 67.71\% \\
9 & 65.99\% \\
\hdashline
AVG & 66.94\% \\
\hline
\multicolumn{2}{c}{$p_2$} \\
\hdashline
0 & 66.91\% \\
1 & 66.75\% \\
2 & 67.5\% \\
3 & 67.34\% \\
4 & 66.83\% \\
5 & 66.62\% \\
6 & 67.82\% \\
7 & 66.43\% \\
8 & 67.21\% \\
9 & 67.7\% \\
\hdashline
AVG & 67.11\% \\
\hline
\multicolumn{2}{c}{Ours} \\
\hdashline
0 & 67.44\% \\
1 & 67.3\% \\
2 & 67.32\% \\
3 & 67.16\% \\
4 & 67.75\% \\
5 & 69.05\% \\
6 & 68.57\% \\
7 & 67.85\% \\
8 & 67.6\% \\
9 & 68.46\% \\
\hdashline
AVG & 67.85\% \\
\hline
% \bottomrule[1.2pt]
\end{tabularx}
\caption{Detailed results for Code Generation Task under Top-$k$ decoding strategy.}
\label{table:appendix_study_decoding_code}
\end{table}

\clearpage
\onecolumn
\subsection{Additional Prompts for Study on Prompt Count $n$}
\label{sec:appendix_study_additional_prompts}

Additional Prompts for Study on Prompt Count $n$ used in Section \ref{sec:study_n} are as following in Table \ref{table:appendix_addtional_prompt_code_example1} and Table \ref{table:appendix_addtional_prompt_ts}:
\begin{table*}[htbp]
\centering
\begin{tabularx}{\textwidth}{X}
\hline
\multicolumn{1}{c}{$p_3$} \\
\hdashline
\textbf{""" This is a piece of code written by an expert that is very ingenious.} \\
def specialFilter(nums): \\
\ \ \ \ \ """Write a function that takes an array of numbers as input and returns the number of elements in the array that are greater than 10 and both first and last digits of a number are odd (1, 3, 5, 7, 9). \\
\ \ \ \ \ For example: \\
\ \ \ \ \ specialFilter([15, -73, 14, -15]) => 1 \\
\ \ \ \ \ specialFilter([33, -2, -3, 45, 21, 109]) => 2 \\
\ \ \ \ \ """ \\
\hline
\multicolumn{1}{c}{$p_4$} \\
\hdashline
\textbf{""" Author: a coder expert} \\
def specialFilter(nums): \\
\ \ \ \ \ """Write a function that takes an array of numbers as input and returns the number of elements in the array that are greater than 10 and both first and last digits of a number are odd (1, 3, 5, 7, 9). \\
\ \ \ \ \ For example: \\
\ \ \ \ \ specialFilter([15, -73, 14, -15]) => 1 \\
\ \ \ \ \ specialFilter([33, -2, -3, 45, 21, 109]) => 2 \\
\ \ \ \ \ """ \\
\hline
\end{tabularx}
\caption{Additional Prompts of ID 146 for Code Generation Task}
\label{table:appendix_addtional_prompt_code_example1}
\end{table*}

\begin{table*}[!h]
\centering
\begin{tabularx}{\textwidth}{X}
\hline
\multicolumn{1}{c}{$p_3$} \\
\hdashline
\textit{\textbf{User Prompt:}} \\
Rewrite this sentence in a simple and easy to understand way. Make sure to retain the meaning and ideas of the original sentence while using shorter words and sentences.\\
Original: \{\textit{input text}\} \\
\hline
\multicolumn{1}{c}{$p_4$} \\
\hdashline
\textit{\textbf{User Prompt:}} \\
Express this sentence in simpler terms, keeping its meaning and ideas intact, and use shorter words and sentences. \\
Original: \{\textit{input text}\} \\
\hline
\end{tabularx}
\caption{Additional Prompts for Text Simplification Task}
\label{table:appendix_addtional_prompt_ts}
\end{table*}

\twocolumn
\end{document}